\pdfoutput=1

\documentclass[11pt]{article}

\usepackage[preprint]{acl}

\usepackage{times}
\usepackage{latexsym}

\usepackage{graphicx}
\usepackage{wrapfig}
\usepackage{amssymb}
\usepackage{amsmath}
\usepackage{bm}
\usepackage{caption}
\usepackage{multirow}
\usepackage{mathrsfs}
\usepackage{mathtools}
\usepackage{algorithm}
\usepackage{algpseudocode}
\usepackage{soul}
\usepackage{booktabs}
\usepackage{tcolorbox}

\usepackage[T1]{fontenc}

\usepackage[utf8]{inputenc}

\usepackage{microtype}

\usepackage{inconsolata}

\usepackage{graphicx}

\title{MEVER: Multi-Modal and Explainable Claim Verification with Graph-based Evidence Retrieval}

\author{
 \textbf{Delvin Ce Zhang\textsuperscript{1}},
 \textbf{Suhan Cui\textsuperscript{2}},
 \textbf{Zhelin Chu\textsuperscript{3}},
 \textbf{Xianren Zhang\textsuperscript{4}},
 \textbf{Dongwon Lee\textsuperscript{5}}
\\
\\
 \textsuperscript{1}University of Sheffield,
 \textsuperscript{2}University of Science and Technology Beijing,\\
 \textsuperscript{3}University of California San Diego,
 \textsuperscript{4,5}The Pennsylvania State University
\\
\\
\textsuperscript{1}\texttt{delvin.ce.zhang@sheffield.ac.uk},
\textsuperscript{2}\texttt{suhan@ustb.edu.cn},\\
\textsuperscript{3}\texttt{12111105@mail.sustech.edu.cn},
\textsuperscript{4,5}\texttt{\{xzz5508,dongwon\}@psu.edu}
}

\begin{document}
\maketitle
\begin{abstract}
Verifying the truthfulness of claims usually requires joint multi-modal reasoning over both textual and visual evidence, such as analyzing both textual caption and chart image for claim verification. In addition, to make the reasoning process transparent, a textual explanation is necessary to justify the verification result. However, most claim verification works mainly focus on the reasoning over textual evidence only or ignore the explainability, resulting in inaccurate and unconvincing verification. To address this problem, we propose a novel model that jointly achieves evidence retrieval, multi-modal claim verification, and explanation generation. For evidence retrieval, we construct a two-layer multi-modal graph for claims and evidence, where we design image-to-text and text-to-image reasoning for multi-modal retrieval. For claim verification, we propose token- and evidence-level fusion to integrate claim and evidence embeddings for multi-modal verification. For explanation generation, we introduce multi-modal Fusion-in-Decoder for explainability. Finally, since almost all the datasets are in general domain, we create a scientific dataset, AIChartClaim, in AI domain to complement claim verification community. Experiments show the strength of our model. Code and datasets are available at \url{https://github.com/cezhang01/mever}.
\end{abstract}

\section{Introduction}

\begin{figure}[t]
	\centering
	\includegraphics[width=1\linewidth]{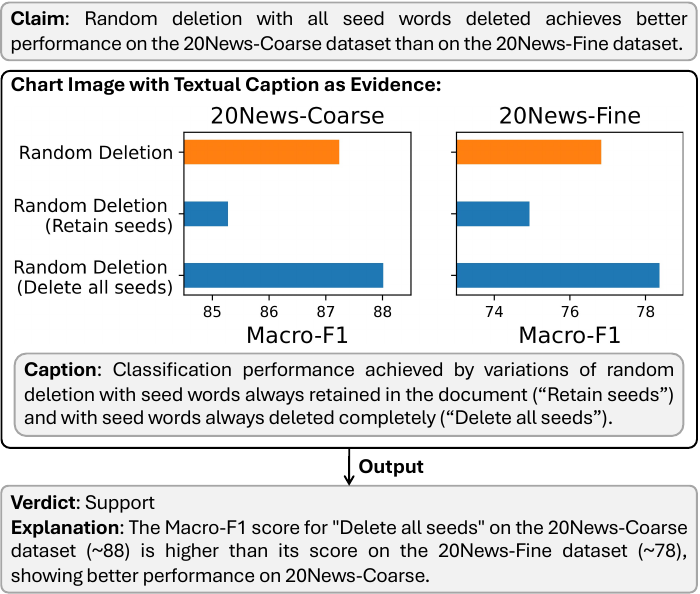}
	\caption{Illustration of multi-modal and explainable claim verification, taken from AIChartClaim dataset.}
	\label{fig:illustration}
\end{figure}

The dissemination of erroneous claims and findings can mislead researchers and the public, leading to unnecessary concerns. This underscores the urgent need for developing a claim verification method to automatically assess the truthfulness of the claims.

Existing methods primarily rely on textual evidence for claim verification. However, with the advent of multimedia, many claims are derived from various types of data, and simply reasoning over textual evidence is insufficient for accurate verification. For example, Fig. \ref{fig:illustration} illustrates a scientific claim concluded from chart images with textual caption. To verify the claim, we need to unify both textual and visual evidence for reasoning and verification. Meanwhile, to make the reasoning process transparent, an additional textual explanation is necessary to justify the ruling process and verification result. However, most existing works mainly focus on the reasoning over textual evidence only \cite{kgat} or ignore the explainability of the ruling process \cite{escnet}, leading to inaccurate and unconvincing verification.

\textbf{Challenges and Approach.} To overcome these limitations, we propose MEVER, \ul{\textbf{M}}ulti-Modal and \ul{\textbf{E}}xplainable Claim \ul{\textbf{V}}erification with \ul{\textbf{E}}vidence \ul{\textbf{R}}etrieval, to address below three open questions.

First, \emph{how to integrate both textual and visual evidence for multi-modal evidence retrieval?} Some works, e.g., MultiVerS \cite{multivers} and DECKER \cite{decker}, rely on external tools \cite{bm25} for evidence retrieval. Other works, e.g., JustiLM \cite{justilm} and RAV \cite{rav}, design an in-built text-only retriever. However, they are uni-modal and ignore visual evidence, leading to inaccurate retrieval. We design a multi-modal evidence retriever. We construct a two-layer multi-modal graph for claims and evidence. Image-to-text and text-to-image reasoning integrates multi-modal data into claim and evidence embeddings for retrieval.

Second, \emph{how to reason between multi-modal claims and evidence for multi-modal verification?} Some models, e.g., GEAR \cite{gear} and Transformer-XH \cite{transformer_xh}, integrate textual claims and evidence for verification, failing to capture visual signals. Recently, multi-modal methods are proposed, e.g., Mocheg \cite{mocheg} and ESCNet \cite{escnet}. However, these methods integrate claims with textual and visual evidence separately, failing to aggregate both textual and visual evidence for joint reasoning. As shown in Fig. \ref{fig:illustration}, images and texts provide complementary information, and jointly aggregating both could reveal insightful discovery. In our model, we design token-level multi-modal fusion and evidence-level hierarchical fusion to achieve fine-grained cross-modal information exchange between claims and evidence. Eventually, we obtain unified claim and evidence embeddings for multi-modal verification.

Third, \emph{how to leverage multi-modal data for explanation generation?} Most explainable models leverage textual claims and evidence to generate explanation, e.g., JustiLM \cite{justilm} and Mocheg \cite{mocheg}. We design a multi-modal Fusion-in-Decoder module, which generates explanation by capturing both multi-modal and multiple pieces of evidence. Besides, to make the explanation consistent with the verification result, we further design a consistency regularizer.

Besides, since most multi-modal datasets are in general domain, we create a scientific dataset in AI domain to complement research community. Our dataset, AIChartClaim in Fig. \ref{fig:illustration}, contains scientific discoveries as claims, chart images with textual captions as evidence, and explanations. It is necessary to have such scientific dataset, since understanding increases and decreases in quantities in charts with scientific language is a crucial reasoning ability for language models. Though ChartCheck \cite{chartcheck} and ChartFC \cite{chartbert} are chart datasets, their content is still in general domain and does not discuss scientific concepts. See Table \ref{table:dataset_comparison} for comparison to selected datasets.

\textbf{Contributions.} \emph{First}, we propose a novel model, MEVER, consisting of multi-modal evidence retrieval, claim verification, and explanation generation. Specifically, we design a two-layer multi-modal graph for evidence retrieval. \emph{Second}, to fully exchange multi-modal information between claims and evidence for verification, we design token- and evidence-level fusion. \emph{Third}, to make the ruling process transparent, we design multi-modal Fusion-in-Decoder and a consistency regularizer for explanation generation. \emph{Fourth}, we create a scientific dataset to complement research community. Experiments verify the strength of our model. 
\section{Related Work}

\textbf{Claim verification.} Previous works are text-only, GEAR \cite{gear}, KGAT \cite{kgat}, TransformerXH \cite{transformer_xh}, MultiVerS \cite{multivers}, HESM \cite{hesm}, DREAM \cite{dream}, CausalWalk \cite{causalwalk}, ProgramFC \cite{program_fc}, CareerScape \cite{fake_resume}, UKE \cite{uke}, and AKEW \cite{akew}. They ignore visual evidence, leading to inaccurate result. Multi-modal methods include Mocheg \cite{mocheg}, ESCNet \cite{escnet}, CCN \cite{ccn}, MR2Retrived \cite{mr2}, CutPaste\&Find \cite{cutpaste}, etc. These methods, except Mocheg, focus on verification, and ignore evidence retrieval or explainability. Our model consists of evidence retrieval, claim verification, and explanation generation.

\textbf{Retrieval-augmented verification} conducts evidence retrieval and claim verification jointly with Retrieval-Augmented Generation \cite{atlas}, such as JustiLM \cite{justilm}, RAV \cite{rav}, RAFTS \cite{rafts}, ARSJoint \cite{arsjoint}, etc. They consider textual evidence only. We construct a two-layer graph for multi-modal evidence retrieval and claim verification.

\textbf{Explainable claim verification} produces textual explanation to justify the verification result \cite{explainable1,explainable2,explainable3,explainable4,mocheg}. They are text-only and do not capture images. Our model designs a multi-modal Fusion-in-Decoder to incorporate multi-modal data for explainability.

\textbf{Scientific and chart claim verification} trains models on scientific and chart claims. SciFact \cite{scifact} is a scientific text-based dataset with no images or explanation. ChartCheck \cite{chartcheck} and ChartFC \cite{chartbert} are chart datasets in general domain. Ours is a multi-modal dataset with explanation in scientific domain. Table \ref{table:dataset_comparison} shows comparison to selected data \cite{fever,feverous,pubhealth,bear_fact,check_covid,factify,newsclippings,mr2,mmcv}. Existing survey \cite{chart_survey} summarizes more chart works, e.g., ChartT5 \cite{chartt5}, MatCha \cite{matcha}, MMC \cite{mmc}, ChartAssistant \cite{chartassistant}.

\textbf{Graph learning} aims to capture both graph structure and node attributes for graph-based tasks \cite{graphsage,adjacent_encoder,dbn,hgtm,scene_graph,hypergraphrag}. However, most of them are designed for self-supervised learning or for supervised generation. Our work focuses on using graph structure to assist multi-modal claim verification.

\begin{table}
	\centering
	\caption{Comparison between selected datasets.}
	\resizebox{\columnwidth}{!}{
		\begin{tabular}{c|ccccc}
			\toprule
			Dataset & Multi-Modal & Explainable & Scientific & Source \\
			\hline
                FEVER &  &  &  & Wiki \\
                FEVEROURS &  &  &  & Wiki \\
                PUBHEALTH &  & $ \checkmark $ &  & FACTWeb \\
                SciFact &  &  & $ \checkmark $ & Paper \\
                BearFact &  &  & $ \checkmark $ & Paper \\
                Check-COVID &  &  & $ \checkmark $ & Paper \\
                \hline
                FACTIFY & $ \checkmark $ &  &  & FACTWeb \\
                NewsCLIPpings & $ \checkmark $ &  &  & Internet \\
                MR2 & $ \checkmark $ &  &  & Internet \\
                MMCV & $ \checkmark $ &  &  & MultimodalQA \\
                ChartFC & $ \checkmark $ &  &  & Internet \\
                ChartCheck & $ \checkmark $ & $ \checkmark $ &  & Internet \\
                Mocheg & $ \checkmark $ & $ \checkmark $ &  & FACTWeb \\
                \hline
                AIChartClaim & $ \checkmark $ & $ \checkmark $ & $ \checkmark $ & Paper \\
			\bottomrule
		\end{tabular}
	}
	\label{table:dataset_comparison}
\end{table}
\section{Model Architecture}

\begin{figure*}[t]
	\centering
	\includegraphics[width=1\linewidth]{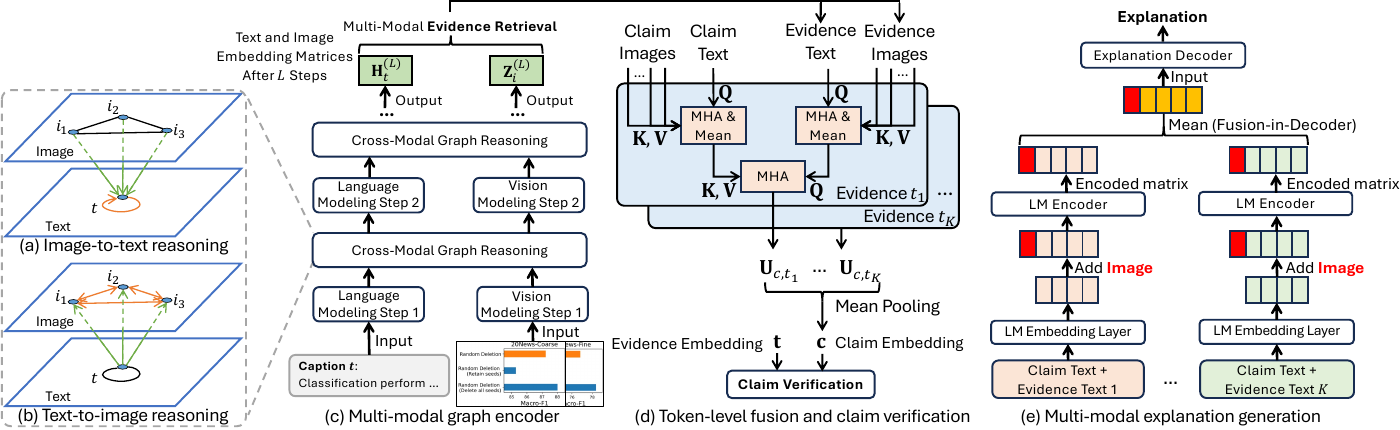}
	\caption{Model architecture. (a-b) Cross-modal graph reasoning. (c) A nested architecture with multi-modal graph reasoning. (d) Multi-modal token-level fusion. (e) Multi-modal explanation generation with Fusion-in-Decoder.}
	\label{fig:model_architecrure}
\end{figure*}

We introduce MEVER, \ul{\textbf{M}}ulti-Modal and \ul{\textbf{E}}xplainable Claim \ul{\textbf{V}}erification with \ul{\textbf{E}}vidence \ul{\textbf{R}}etrieval. Table \ref{table:notation_table} summarizes math notations.

\subsection{Problem Formulation}

We have a multi-modal dataset $ \mathcal{D}=\{\mathcal{C},\mathcal{T},\mathcal{I},\mathcal{E}\} $. $ \mathcal{C}=\{c_n\}_{n=1}^N $ is a set of $ N $ claims. Textual evidence set $ \mathcal{T}=\{t_m\}_{m=1}^T $ is a corpus of $ T $ evidence texts. Each evidence text $ t $ is associated with a set of images $ \mathcal{I}(t)=\{i_t\} \subset \mathcal{I} $ where $ \mathcal{I} $ is an image set. An evidence text may have multiple images, e.g., a caption with multiple charts in Fig. \ref{fig:illustration}. Sometimes, claim $ c $ also has images $ \mathcal{I}(c)=\{i_c\} \subset \mathcal{I} $. $ \mathcal{E}=\{e_n\}_{n=1}^N $ is a set of $ N $ explanations for $ N $ corresponding claims. If we do not observe the association between texts and images, we use pre-trained CLIP \cite{clip} for alignment.

Given $ \mathcal{C} $, $ \mathcal{T} $, and $ \mathcal{I} $ as \ul{inputs}, our model uses textual evidence $ \mathcal{T} $ and visual evidence $ \mathcal{I} $ to verify claims $ \mathcal{C} $. For each claim $ c $, we have two \ul{outputs}. One is claim's veracity label $ \hat{y}\in\mathcal{Y}=\{\text{SUPPORT, REFUTE, NEI}\} $, i.e., whether the evidence supports, refutes, or does not have enough information to verify the claim. The other is explanation $ \hat{e} $ for reasoning process. Note that some datasets \cite{chartcheck} do not have NEI label.

Fig. \ref{fig:model_architecrure} shows the overall model architecture with three main modules, (a-c) evidence retrieval, (d) claim verification, and (e) explanation generation.

\subsection{Evidence Retrieval with Two-Layer Multi-Modal Graph}

\textbf{Graph construction.} For each evidence text $ t \in \mathcal{T} $ and its associated images $ \mathcal{I}(t) \subset \mathcal{I} $, we construct a two-layer multi-modal graph in Fig. \ref{fig:model_architecrure}(a-b). The bottom layer is textual layer, and each node is an evidence text. The top layer is visual layer, and each node is an image. Cross-layer edges denote the correspondence between text $ t $ and its images $ \mathcal{I}(t) $. We also add intra-layer edges. For visual layer, we add fully connected edges among images of the same text for multi-image reasoning. For textual layer, we add self-loop edge. During evidence retrieval, we treat each evidence independently, thus we add self-loop edge. During verification in Sec. \ref{sec:claim_verification}, we will add fully connected edges among multiple retrieved evidence for multi-evidence reasoning.

\textbf{Image-to-text reasoning.} Evidence reasoning consists of image-to-text and text-to-image reasoning. We first present image-to-text reasoning here.

For each evidence text $ t $, we use $ \textbf{H}_t^{(l)}=[\textbf{h}_{t,\text{CLS}}^{(l)},\textbf{h}_{t,1}^{(l)},\textbf{h}_{t,2}^{(l)},...] $ to represent the output from the $ l $-th Transformer step. Here we use ``step'' to replace ``layer'' \cite{transformer} to differentiate from our two-layer graph. Similarly, for each evidence image $ i $, we have output from the $ l $-th ViT step $ \textbf{Z}_i^{(l)}=[\textbf{z}_{i,\text{CLS}}^{(l)},\textbf{z}_{i,1}^{(l)},\textbf{z}_{i,2}^{(l)},...] $. Since an evidence text may have multiple images, we use Graph Neural Network (GNN) \cite{graphsage} to aggregate multiple images. We first project text and image embeddings into the same space.
\begin{equation}
\label{eq:linear_projection}
    \tilde{\textbf{h}}_{t,\text{CLS}}^{(l)}=\textbf{W}_{\text{txt}}\textbf{h}_{t,\text{CLS}}^{(l)},\;\; \tilde{\textbf{z}}_{i,\text{CLS}}^{(l)}=\textbf{W}_{\text{img}}\textbf{z}_{i,\text{CLS}}^{(l)}.
\end{equation}
$ \textbf{W}_{\text{txt}},\textbf{W}_{\text{img}}\in\Bbb R^{d\times d} $ are projection matrices for text and image, respectively. We use [CLS] tokens as text and image embeddings. We design image-to-text attention, shown by green arrows in Fig. \ref{fig:model_architecrure}(a).
\begin{equation}
\label{eq:image_to_text_attention}
\resizebox{\columnwidth}{!}{
	$ a_{t,i}=\text{softmax}\Bigl(\text{sigmoid}(\textbf{b}_{\text{i2t}}^{\top}[\tilde{\textbf{h}}_{t,\text{CLS}}^{(l)}||\tilde{\textbf{z}}_{i,\text{CLS}}^{(l)}])\Bigl) $
 }
\end{equation}
where $ i\in\mathcal{I}(t) $ is an evidence image that text $ t $ is associated with, $ [\cdot||\cdot] $ is concatenation, and $ \textbf{b}_{\text{i2t}}\in\Bbb R^{2d} $. We aggregate images by $ \hat{\textbf{z}}_t^{(l)}=\sum_{i\in\mathcal{I}(t)}a_{t,i}\tilde{\textbf{z}}_{i,\text{CLS}}^{(l)} $. Summarizing above GNN module, we have
\begin{equation}
\label{eq:graph_conv_layer}
\resizebox{\columnwidth}{!}{
    $ \hat{\textbf{z}}_t^{(l)}=f_{\text{GNN}}\Big(\textbf{h}_{t,\text{CLS}}^{(l)},\{\textbf{z}_{i,\text{CLS}}^{(l)}|i\in\mathcal{I}(t)\};\textbf{W}_{\text{txt}},\textbf{W}_{\text{img}},\textbf{b}_{\text{i2t}}\Big). $
}
\end{equation}

Since textual layer has self-loop, we set $ \hat{\textbf{h}}_t^{(l)}=\tilde{\textbf{h}}_{t,\text{CLS}}^{(l)} $ without GNN (orange arrow in Fig. \ref{fig:model_architecrure}(a)).

Finally, we integrate the aggregated image and text embeddings into evidence text for cross-modal reasoning. We consider $ \hat{\textbf{z}}_t^{(l)} $ and $ \hat{\textbf{h}}_t^{(l)} $ as \emph{virtual tokens} and concatenate with $ t $'s embedding matrix by $ \widehat{\textbf{H}}_t^{(l)}=[\hat{\textbf{z}}_t^{(l)}||\hat{\textbf{h}}_t^{(l)}||\textbf{H}_t^{(l)}] $. 
To fully unify multi-modal data, we input $ \widehat{\textbf{H}}_t^{(l)} $ to the $ (l+1) $-th Transformer step with multi-head attention.
\begin{equation}
\label{eq:multi_head_attention}
\resizebox{\columnwidth}{!}{
    $ \text{MHA}(\textbf{Q}=\textbf{W}_Q^{(l)}\textbf{H}_t^{(l)},\textbf{K}=\textbf{W}_K^{(l)}\widehat{\textbf{H}}_t^{(l)},\textbf{V}=\textbf{W}_V^{(l)}\widehat{\textbf{H}}_t^{(l)}). $
}
\end{equation}
Key $ \textbf{K} $ and value $ \textbf{V} $ are augmented with multi-modal virtual tokens. After multi-layer perceptron and layer normalization, we have evidence text $ t $'s embedding matrix $ \textbf{H}_t^{(l+1)} $ at the $ (l+1) $-th step.

\textbf{Text-to-image reasoning.} We symmetrically introduce text-to-image reasoning and propagate text embeddings to visual layer for ViT modeling in Fig. \ref{fig:model_architecrure}(b). Since each image $ i $ is usually associated with one evidence text $ t $, such as a chart with its caption, text-to-image reasoning simply becomes $ \hat{\textbf{h}}_t^{(l)}=\tilde{\textbf{h}}_{t,\text{CLS}}^{(l)}=\textbf{W}_{\text{txt}}\textbf{h}_{t,\text{CLS}}^{(l)} $. For intra-layer multi-image reasoning, we have GNN module.
\begin{equation}
\label{eq:text_to_image_reasoning}
\resizebox{\columnwidth}{!}{
    $ \hat{\textbf{z}}_i^{(l)}=f_{\text{GNN}}\Big(\textbf{z}_{i,\text{CLS}}^{(l)},\{\textbf{z}_{i^\prime,\text{CLS}}^{(l)}|i^\prime\in\mathcal{I}(t)\};\textbf{W}_{\text{img}},\textbf{b}_{\text{i2i}}\Big). $
}
\end{equation}
We consider $ \hat{\textbf{h}}_t^{(l)} $ and $ \hat{\textbf{z}}_i^{(l)} $ as virtual tokens and concatenate with $ \textbf{Z}_i^{(l)} $, i.e., $ \widehat{\textbf{Z}}_i^{(l)}=[\hat{\textbf{z}}_i^{(l)}||\hat{\textbf{h}}_t^{(l)}||\textbf{Z}_i^{(l)}] $. This matrix is input to the $ (l+1) $-th ViT step \cite{vit}, similarly to Eq. \ref{eq:multi_head_attention}.

We repeat image-to-text and text-to-image reasoning inside each Transformer and ViT step for $ L $ steps, and obtain a nested encoder in Fig. \ref{fig:model_architecrure}(c).
\begin{equation}
\label{eq:multi_modal_graph_encoder}
    \textbf{H}^{(L)}_{t},\{\textbf{Z}^{(L)}_{i}\}_{i\in\mathcal{I}(t)}=f_{\text{Enc}}\big(t,\{i|i\in\mathcal{I}(t)\}\big).
\end{equation}
We take $ \textbf{h}_t=\textbf{h}_{t,\text{CLS}}^{(L)} $ as evidence embedding, since it already absorbs both text and image information. For claim $ c $ with its associated images, we input to the same multi-modal encoder and obtain claim embedding $ \textbf{h}_c=\textbf{h}_{c,\text{CLS}}^{(L)} $. We use below contrastive loss as retrieval objective. $ t^\prime $ is a negative evidence in the same mini-batch $ B $.
\begin{equation}
\label{eq:retrieval_objective_function}
\resizebox{\columnwidth}{!}{
    $ \mathcal{L}_{\text{Ret}}=-\sum_{c\in\mathcal{C}_{\text{train}}}\log\dfrac{\exp(\textbf{h}_c^\top\textbf{h}_t)}{\exp(\textbf{h}_c^\top\textbf{h}_t)+\sum_{t^\prime\in B\textbackslash t}\exp(\textbf{h}_{c}^\top\textbf{h}_{t^\prime})}. $
}
\end{equation}

\subsection{Claim Verification with Token- and Evidence-Level Fusion}
\label{sec:claim_verification}

After we obtain the retrieved evidence with texts and associated images, here we present claim verification using multi-modal evidence. Above multi-modal graph encoder outputs text $ \textbf{H}_t=\textbf{H}^{(L)}_t\in\Bbb R^{P_\text{txt}\times d} $ and image embedding matrices $ \textbf{Z}_i=\textbf{Z}^{(L)}_i\in\Bbb R^{P_{\text{img}}\times d} $ for both claims and evidence. $ P_{\text{txt}} $ and $ P_{\text{img}} $ respectively denotes the number of textual tokens and visual patches. We allow claims and evidence to interact with each other for claim verification. 

\textbf{Token-level multi-modal fusion.} We aim to design an interactive fusion to exchange information between claims and evidence. For each claim $ c $, we have $ K $ retrieved evidence texts $ \{t_k\}_{k=1}^K $ and images $ \mathcal{I}(t_k) $ for each text $ t_k $. We design an interactive fusion method between claims and evidence at token level, shown by Fig. \ref{fig:model_architecrure}(d). For evidence text $ t_k $ and each of its associated images $ i\in\mathcal{I}(t_k) $, we leverage a multi-modal multi-head attention for token-level fusion and obtain aggregated image embedding matrix $ \textbf{Z}_{t_k,i}\in\Bbb R^{P_{\text{txt}}\times d} $.
\begin{equation}
\label{eq:multi_head_attention_token_level}
    \textbf{Z}_{t_k,i}=\text{MHA}(\textbf{W}_Q\textbf{H}_{t_k},\textbf{W}_K\textbf{Z}_i,\textbf{W}_V\textbf{Z}_i).
\end{equation}
Since an evidence text $ t_k $ has multiple images, we obtain a set $ \{\textbf{Z}_{t_k,i}\}_{i\in\mathcal{I}(t_k)} $. We take mean pooling to obtain a single image embedding matrix by
\begin{equation}
    \textbf{Z}_{t_k}=\text{mean}\big(\{\textbf{Z}_{t_k,i}|i\in\mathcal{I}(t_k)\}\big).
\end{equation}
Finally, we unify both text embedding matrix $ \textbf{H}_{t_k} $ and image embedding matrix $ \textbf{Z}_{t_k} $ by
\begin{equation}
\label{eq:linear_projection_token_level}
    \textbf{U}_{t_k}=\textbf{W}_1[\textbf{H}_{t_k}||\textbf{Z}_{t_k}].
\end{equation}
$ \textbf{W}_1\in\Bbb R^{d\times 2d} $, and $ \textbf{U}_{t_k}\in\Bbb R^{P_{\text{txt}}\times d} $ is the unified multi-modal matrix for evidence $ t_k $. For claim $ c $ and its images $ \mathcal{I}(c) $, we repeat Eqs. \ref{eq:multi_head_attention_token_level}--\ref{eq:linear_projection_token_level} to obtain unified multi-modal matrix for claim $ \textbf{U}_c\in\Bbb R^{P_{\text{txt}}\times d} $.

To allow claim-evidence interaction, we have
\begin{equation}
\label{eq:claim_evidence_interaction}
    \textbf{U}_{t_k,c}=\text{MHA}(\textbf{W}_Q\textbf{U}_{t_k},\textbf{W}_K\textbf{U}_c,\textbf{W}_V\textbf{U}_c).
\end{equation}
Here $ \textbf{U}_{t_k,c} $ integrates claim $ c $ and evidence $ t_k $ as well as their associated images. Since a claim has multiple retrieved evidence, we take mean pooling and obtain a single claim embedding matrix $ \textbf{U}_c=\text{mean}\big(\{\textbf{U}_{t_k,c}|t_k\in\mathcal{T}(c)\}\big) $ where $ \mathcal{T}(c) $ is a set of retrieved evidence. Finally, we take [CLS] in $ \textbf{U}_c $ as claim $ c $'s embedding $ \textbf{c}=\textbf{u}_{c,\text{CLS}} $, which fuses claim and evidence at token level. See Fig. \ref{fig:model_architecrure}(d).

\textbf{Evidence-level hierarchical fusion.} We now present evidence embedding. A claim has $ K $ retrieved evidence texts, and each text is associated with $ |\mathcal{I}(t_k)| $ images, resulting in a hierarchical structure. We propose a hierarchical fusion to obtain multi-modal evidence embedding. Specifically, we have multi-modal text $ \textbf{H}_t=\textbf{H}^{(L)}_t $ and image embeddings $ \textbf{Z}_i=\textbf{Z}^{(L)}_i $, output from multi-modal graph encoder in Eq. \ref{eq:multi_modal_graph_encoder}. For each evidence text $ t_k\in\mathcal{T}(c) $, we use GNN to aggregate its images.
\begin{equation}
\label{eq:image_fusion}
\resizebox{\columnwidth}{!}{
    $ \hat{\textbf{z}}_{t_k}=f_{\text{GNN}}\Big(\textbf{h}_{t_k,\text{CLS}},\{\textbf{z}_{i,\text{CLS}}|i\in\mathcal{I}(t_k)\};\textbf{W}_{\text{txt}},\textbf{W}_{\text{img}},\textbf{b}_{\text{i2t}}\Big). $
}
\end{equation}
The aggregated image embedding is then combined with evidence text embedding by $ \textbf{t}_{k}=\textbf{W}_2[\textbf{h}_{t_k,\text{CLS}}||\hat{\textbf{z}}_{t_k}] $ where $ \textbf{W}_2\in\Bbb R^{d\times 2d} $ and $ \textbf{t}_{k} $ is evidence embedding. Finally, we use claim embedding $ \textbf{c} $ as query to aggregate evidence embeddings.
\begin{equation}
\label{eq:evidence_fusion}
\resizebox{\columnwidth}{!}{
    $ \textbf{t}=f_{\text{GNN}}\Big(\textbf{c},\{\textbf{t}_{k}|t_k\in\mathcal{T}(c)\};\textbf{W}_{\text{txt}},\textbf{W}_{\text{evid}},\textbf{b}_{\text{e2c}}\Big). $
}
\end{equation}

Having obtained claim and evidence embeddings, we input them into a classifier for verification, $ \hat{\textbf{y}}= \text{softmax}(f_{\text{MLP}}([\textbf{c}||\textbf{t}])) $ with below loss.
\begin{equation}
\label{eq:verification_loss}
    \mathcal{L}_{\text{Ver}}=-\sum_{y^\prime\in\mathcal{Y}}y^\prime\log \hat{y}^\prime.
\end{equation}

\subsection{Multi-Modal Explanation Generation}

We use multi-modal data to generate explanations to justify the ruling process. We design Fusion-in-Decoder (Fig. \ref{fig:model_architecrure}(e)) and consistency regularizer.

\textbf{Multi-modal Fusion-in-Decoder.} We are given a claim text $ c $ and each of its evidence texts $ t_k\in\mathcal{T}(c) $, we concatenate their raw texts into a single textual sequence with [SEP] token as separator, i.e., $ [c||[SEP]||t_k] $. In language models, there is an embedding look-up table before language encoder. Textual sequence is first mapped to token embeddings in this look-up table, which are then summed up with positional encodings. We thus obtain embedding matrix $ \textbf{E}_{c,t_k} $ for the concatenated sequence using this look-up table. We then concatenate the associated image embeddings of claim and evidence text to obtain a multi-modal matrix.
\begin{equation}
\label{eq:concatenated_embedding_matrix}
\resizebox{\columnwidth}{!}{
    $ \widehat{\textbf{E}}_{c,t_k}=[\underbrace{\textbf{W}_{\text{img}}\textbf{z}_{c,\text{CLS}}||...}_{\text{claim's images}}||\underbrace{\textbf{W}_{\text{img}}\textbf{z}_{t_k,\text{CLS}}||...}_{\text{evidence text's images}}||\textbf{E}_{c,t_k}] $
}
\end{equation}
$ \textbf{z}_{c,\text{CLS}} $ and $ \textbf{z}_{t_k,\text{CLS}} $ are respectively claim $ c $'s and evidence text $ t_k $'s image embeddings, obtained from multi-modal graph encoder in Eq. \ref{eq:multi_modal_graph_encoder}. We project them using $ \textbf{W}_{\text{img}} $ to the same embedding space as texts. We input the concatenated multi-modal embedding matrix $ \widehat{\textbf{E}}_{c,t_k} $ to language encoder, T5 \cite{t5}, and obtain $ \tilde{\textbf{E}}_{c,t_k}=f_{\text{LMEnc}}(\widehat{\textbf{E}}_{c,t_k}) $.

Since a claim $ c $ has multiple retrieved evidence $ \mathcal{T}(c) $, we design multi-modal Fusion-in-Decoder module to capture all evidence for explanation generation, i.e., $ \hat{e}=f_{\text{LMDec}}(\bar{\textbf{E}}_{c,t}) $ where $ \bar{\textbf{E}}_{c}=\text{mean}(\tilde{\textbf{E}}_{c,t_1},...,\tilde{\textbf{E}}_{c,t_K}) $.
This module is illustrated by Fig. \ref{fig:model_architecrure}(e). We have generation loss below.
\begin{equation}
\label{eq:generation_loss}
    \mathcal{L}_{\text{Exp}}=-\sum_{e_j}\log p(e_j|e_{<j},\bar{\textbf{E}}_{c})
\end{equation}
$ e_{<j} $ denotes the tokens generated prior to $ e_j $, and $ p(e_j|\cdot)=\text{softmax}(\bm{\mathscr{L}}(e_j|\cdot)) $ is the probability of $ e_j $, calculated by normalizing logits $ \bm{\mathscr{L}}(e_j|\cdot) $.

\textbf{Consistency regularizer.} The generated explanation should consistently justify the predicted verification. Thus, we design a consistency regularizer to achieve this goal. Specifically, we do mean pooling for logits of all the tokens in explanation by
\begin{equation}
\label{eq:logits_mean_pooling}
    \bm{\mathscr{L}}=\text{mean}(\bm{\mathscr{L}}(e_1|\cdot),...,\bm{\mathscr{L}}(e_j|\cdot),...).
\end{equation}
$ \bm{\mathscr{L}}\in\Bbb R^{V} $ is a $ V $-dimensional embedding where $ V $ is the length of language model vocabulary. $ \bm{\mathscr{L}} $ contains the information of explanation, and its embedding should also reveal verification label. Thus, we input $ \bm{\mathscr{L}} $ to a classifier to predict verification label by $ \hat{\textbf{y}}_e=\text{softmax}(f_{\text{MLP}}(\bm{\mathscr{L}})) $. Finally, we minimize the difference between the predicted label $ \hat{\textbf{y}} $ of the verification module and $ \hat{\textbf{y}}_e $ using
\begin{equation}
\label{eq:regularization_loss}
    \mathcal{L}_{\text{Reg}}=\text{KL}(\hat{\textbf{y}}||\hat{\textbf{y}}_e)+\text{KL}(\hat{\textbf{y}}_e||\hat{\textbf{y}})-\sum_{y^\prime\in\mathcal{Y}}y^\prime\log \hat{y}_e^\prime.
\end{equation}
We sum up two KL divergences \cite{prml} to remove its asymmetry. We also add a cross-entropy loss for the predicted label of logits. 
The overall loss function becomes
\begin{equation}
\label{eq:ver_and_exp_loss}
    \mathcal{L}=\mathcal{L}_{\text{Ver}}+\mathcal{L}_{\text{Exp}}+\lambda\mathcal{L}_{\text{Reg}}.
\end{equation}
$ \lambda=0.5 $ is a hyperparameter for the importance of regularizer. 
We summarize the learning in Algos. \ref{algo:training_algorithm1}--\ref{algo:training_algorithm2}. See Appendix \ref{sec:complexity_analysis} for complexity analysis.
\section{Dataset Creation}
\label{sec:dataset_creation}

We create a scientific dataset. We present main creation here and put more details in Appendix \ref{sec:additional_dataset_details}.

\textbf{Data source.} We use AI papers as source. Other domains are future work. We have 5 categories, and each category has 3 conferences, AI (AAAI, IJCAI, UAI), ML (NeurIPS, ICML, ICLR), NLP (ACL, EMNLP, NAACL), CV (CVPR, ICCV, ECCV), Data Mining (KDD, WWW, WSDM), i.e., totally 15 conferences. For each conference, we collect recent 4 proceedings. For each proceeding, we select 5 papers with chart, with totally 300 papers.

Within each paper, we collect its chart with caption as multi-modal evidence. The sentences in the main text that mention the chart usually contain scientific discoveries or claims. However, not every claim is checkworthy and not every chart is clear. To ensure the quality, our annotators (one postdoc and three PhD students specializing in AI) filter out inappropriate claims and charts and search for other papers with high-quality data. Finally, we have 300 claims with corresponding 300 charts and captions. These charts include line, bar, pie, scatterplot, and heat map. See Table \ref{table:ai_chart_claim_dataset_statistics} for details.

\textbf{Data augmentation.} Usually, the charts support the claims, thus above 300 claims, undergoing manual double check by annotators, are labeled as ``SUPPORT''. To create claims refuted by the charts, we follow \citet{scifact} and ask annotators to write negation for the 300 claims, taking precautions not to bias the negation by using obvious keywords, like ``not''. We finally obtain 600 claims, half supported and half refuted by the charts. 

To augment the dataset, we follow \citet{averitec} and use GPT-4o \cite{gpt4} to generate 600 more claims. For each of 300 charts with captions, we use the prompt in Appendix \ref{sec:additional_dataset_details} to generate two more claims, one supported and one refuted by the chart. After generation, we have 1,200 claims, i.e., 600 natural and 600 generated claims. Our annotators rigorously check the generation and make manual corrections when necessary to ensure the dataset is correct and high-quality.

\textbf{Explanation.} A significant portion of papers do not have explanations for claims. Moreover, we do not have explanations for the generated claims. 
For consistency, we use GPT-4o to generate explanations with prompt in Appendix \ref{sec:additional_dataset_details}. The generation is rigorously checked and corrected by annotators if there is erroneous or unclear description.

\textbf{Size.} Finally, we have 1,200 claims with explanations and 300 charts with captions. The size is comparable to textual scientific datasets, e.g., 1,409 claims in SciFact \cite{scifact}, 1,448 claims in BearFact \cite{bear_fact}, 1,504 claims in Check-COVID \cite{check_covid}. There are two constraints for the size. For one, the number of AI papers with checkworthy claims and readable charts is limited. Our annotators have tried hard to obtain appropriate papers. For the other, the creation requires experts to manually analyze the charts. The number of domain experts is limited. We consider creating larger-size dataset as future work. Also, we follow \citet{chartcheck} and do not include NEI label for clarity and consistency. We can also add NEI by simply replacing the gold evidence of each claim with random evidence.
\section{Experiments}

\begin{table}
	\centering
	\caption{Dataset statistics.}
	\resizebox{\columnwidth}{!}{
		\begin{tabular}{c|cccc}
			\toprule
			Name & \#Claims & \#Evidence Texts & \#Images & Explanation \\
                \hline
                AIChartClaim & 1,200 & 300 & 300 & Yes \\
                ChartCheck & 10,038 & 1,615 & 1,615 & Yes \\
                Mocheg & 15,601 & 33,880 & 13,052 & Yes \\
                MR2 & 13,785 & 91,347 & 105,132 & No \\
			\bottomrule
		\end{tabular}
	}
	\label{table:dataset_statistics}
\end{table}

\begin{table*}[t]
	\centering
	\caption{Result of evidence retrieval (\%). Standard deviation of BM25 is 0, because it is a deterministic approach.}
	\resizebox{\textwidth}{!}{
		\begin{tabular}{c|ccc|ccc|ccc|ccc}
			\toprule
                \multirow{2}{*}{Model} & \multicolumn{3}{c|}{AIChartClaim} & \multicolumn{3}{c|}{ChartCheck} & \multicolumn{3}{c|}{Mocheg} & \multicolumn{3}{c}{MR2} \\
			\cline{2-13}
			{} & MAP & Prec@3 & Rec@3 & MAP & Prec@3 & Rec@3 & MAP & Prec@3 & Rec@3 & MAP & Prec@3 & Rec@3 \\
			\hline
                BM25 & 49.5$ \pm $0.0 & 17.6$ \pm $0.0 & 52.9$ \pm $0.0 & 40.6$ \pm $0.0 & 14.6$ \pm $0.0 & 43.6$ \pm $0.0 & 27.3$ \pm $0.0 & 21.2$ \pm $0.0 & 26.8$ \pm $0.0 & 11.7$ \pm $0.0 & 17.5$ \pm $0.0 & 8.4$ \pm $0.0 \\
                RAV & 59.0$ \pm $0.8 & 21.6$ \pm $0.3 & 64.7$ \pm $1.0 & 59.9$ \pm $0.2 & 21.6$ \pm $0.2 & 64.8$ \pm $0.7 & 39.0$ \pm $0.8 & 29.7$ \pm $0.6 & 37.6$ \pm $0.8 & 15.2$ \pm $0.3 & 22.1$ \pm $0.6 & 10.7$ \pm $0.4 \\
                JustiLM & 62.0$ \pm $1.2 & 22.4$ \pm $1.0 & 64.2$ \pm $3.0 & 58.3$ \pm $0.5 & 21.1$ \pm $0.2 & 63.4$ \pm $0.7 & 39.5$ \pm $0.4 & 29.8$ \pm $0.2 & 38.0$ \pm $0.2 & 14.5$ \pm $0.2 & 20.7$ \pm $0.5 & 10.1$ \pm $0.2 \\
                \hline
                MochegModel & 65.9$ \pm $0.0 & 23.3$ \pm $0.0 & 70.0$ \pm $0.0 & 58.1$ \pm $0.0 & 21.0$ \pm $0.0 & 63.0$ \pm $0.1 & 36.0$ \pm $0.0 & 27.9$ \pm $0.0 & 38.2$ \pm $0.0 & 16.7$ \pm $0.0 & 23.2$ \pm $0.0 & 11.2$ \pm $0.0 \\
                TransXH+ViT & 62.0$ \pm $2.5 & 22.7$ \pm $0.7 & 68.1$ \pm $2.1 & 60.8$ \pm $0.3 & 21.6$ \pm $0.1 & 64.8$ \pm $0.3 & 39.8$ \pm $0.5 & 30.2$ \pm $0.3 & 38.1$ \pm $0.4 & 17.8$ \pm $0.6 & 28.2$ \pm $0.3 & 12.6$ \pm $0.5 \\
                \hline
                MEVER w/o images & 69.8$ \pm $1.0 & 24.9$ \pm $0.5 & 74.7$ \pm $1.5 & 58.3$ \pm $0.9 & 21.2$ \pm $0.3 & 63.7$ \pm $0.9 & 38.6$ \pm $2.9 & 29.3$ \pm $2.5 & 37.3$ \pm $2.9 & 15.0$ \pm $0.2 & 21.7$ \pm $0.4 & 10.6$ \pm $0.2 \\
                \hline
                MEVER (ours) & \textbf{71.4$ \pm $0.3} & \textbf{25.4$ \pm $0.1} & \textbf{76.1$ \pm $0.2} & \textbf{63.6$ \pm $0.3} & \textbf{22.7$ \pm $0.1} & \textbf{68.0$ \pm $0.3} & \textbf{41.6$ \pm $0.4} & \textbf{31.7$ \pm $0.3} & \textbf{40.1$ \pm $0.2} & \textbf{19.5$ \pm $0.4} & \textbf{29.8$ \pm $0.4} & \textbf{13.1$ \pm $0.3} \\
			\bottomrule
		\end{tabular}
	}
	\label{table:evidence_retrieval}
\end{table*}

\begin{table*}[t]
	\centering
	\caption{Claim verification with \emph{Macro F1} score (\%).}
	\resizebox{\textwidth}{!}{
		\begin{tabular}{c|cc|cc|cc|cc}
			\toprule
                \multirow{2}{*}{Model} & \multicolumn{2}{c|}{AIChartClaim} & \multicolumn{2}{c|}{ChartCheck} & \multicolumn{2}{c|}{Mocheg} & \multicolumn{2}{c}{MR2} \\
			\cline{2-9}
			{} & Gold & Retrieved & Gold & Retrieved & Gold & Retrieved & Gold & Retrieved \\
			\hline
                KGAT & 68.7$ \pm $0.4 & 68.8$ \pm $0.3 & 62.4$ \pm $0.2 & 60.4$ \pm $0.3 & 47.4$ \pm $0.6 & 45.1$ \pm $0.5 & 65.0$ \pm $1.5 & 64.3$ \pm $1.1 \\
                CausalWalk & 68.4$ \pm $2.3 & 69.0$ \pm $2.4 & 60.4$ \pm $2.8 & 61.3$ \pm $0.9 & 38.6$ \pm $0.7 & 30.1$ \pm $0.6 & 64.6$ \pm $2.5 & 65.3$ \pm $4.8 \\
                CORRECT & 67.9$ \pm $1.5 & 70.0$ \pm $1.2 & 62.3$ \pm $0.4 & 61.5$ \pm $0.3 & 46.0$ \pm $0.9 & 45.9$ \pm $1.7 & 70.0$ \pm $1.5 & 70.4$ \pm $3.0 \\
                TransformerXH & 68.6$ \pm $0.2 & 69.1$ \pm $0.6 & 62.9$ \pm $0.4 & 62.0$ \pm $0.9 & 43.8$ \pm $0.9 & 44.3$ \pm $0.6 & 68.1$ \pm $1.2 & 70.9$ \pm $1.7 \\
                \hline
                TransformerXH++ & 69.4$ \pm $0.9 & 68.3$ \pm $1.0 & 63.2$ \pm $0.2 & 62.6$ \pm $0.6 & 45.0$ \pm $1.4 & 44.5$ \pm $0.1 & 69.6$ \pm $0.4 & 72.5$ \pm $1.0 \\
                MochegModel & 57.9$ \pm $1.7 & 59.0$ \pm $2.4 & 56.7$ \pm $1.6 & 57.8$ \pm $1.3 & 45.6$ \pm $1.4 & 45.5$ \pm $1.2 & 64.8$ \pm $0.3 & 68.0$ \pm $0.4 \\
                MR2Retrieved & 69.9$ \pm $0.5 & 69.8$ \pm $0.2 & 62.7$ \pm $0.1 & 62.3$ \pm $0.9 & 44.0$ \pm $0.2 & 44.5$ \pm $0.6 & 73.9$ \pm $1.2 & 71.2$ \pm $0.8 \\
                CCN & 69.8$ \pm $0.9 & 69.3$ \pm $0.2 & 62.6$ \pm $0.4 & 62.6$ \pm $0.5 & 44.9$ \pm $0.4 & 47.5$ \pm $0.4 & 73.5$ \pm $1.3 & 75.5$ \pm $2.1 \\
                ESCNet & 69.5$ \pm $0.6 & 69.6$ \pm $0.4 & 60.7$ \pm $0.5 & 61.3$ \pm $0.5 & 46.4$ \pm $0.3 & 47.4$ \pm $0.5 & 73.5$ \pm $0.6 & 75.0$ \pm $0.9 \\
                ECENet & 70.0$ \pm $0.7 & 70.2$ \pm $0.7 & 60.9$ \pm $0.8 & 61.3$ \pm $0.9 & 45.5$ \pm $1.2 & 46.7$ \pm $1.8 & 72.4$ \pm $1.2 & 74.2$ \pm $0.8 \\
                MultiKE-GAT & 66.6$ \pm $0.4 & 67.3$ \pm $0.5 & 60.5$ \pm $0.5 & 60.6$ \pm $0.8 & 39.9$ \pm $1.4 & 46.2$ \pm $1.0 & 67.0$ \pm $1.8 & 71.6$ \pm $2.0 \\
                GPT-4o & 51.0$ \pm $1.7 & 41.7$ \pm $1.2 & 43.9$ \pm $0.8 & 49.8$ \pm $1.8 & \textbf{48.7$ \pm $3.7} & 44.5$ \pm $3.5 & 68.7$ \pm $2.4 & 63.7$ \pm $2.8 \\
                \hline
                ChartBERT & 42.6$ \pm $4.7 & 43.0$ \pm $1.6 & 55.7$ \pm $0.6 & 40.9$ \pm $2.9 & N.A. & N.A. & N.A. & N.A.\\
                UniChart & 69.3$ \pm $1.6 & 68.4$ \pm $0.4 & 62.6$ \pm $0.6 & 62.3$ \pm $0.1 & N.A. & N.A. & N.A. & N.A.\\
                ChartGemma & 69.1$ \pm $0.4 & 68.6$ \pm $1.4 & 63.3$ \pm $0.3 & 63.9$ \pm $0.3 & N.A. & N.A. & N.A. & N.A.\\
                \hline
                JustiLM & 65.4$ \pm $0.7 & 65.3$ \pm $0.2 & 61.3$ \pm $0.5 & 62.3$ \pm $0.1 & 38.5$ \pm $0.7 & 44.4$ \pm $0.4 & 69.7$ \pm $1.1 & 72.1$ \pm $0.5 \\
                RAV & 67.5$ \pm $0.4 & 67.6$ \pm $0.2 & 61.8$ \pm $1.2 & 60.3$ \pm $1.5 & 45.8$ \pm $1.1 & 42.5$ \pm $0.4 & 65.0$ \pm $1.7 & 61.9$ \pm $2.6 \\
                \hline
                MEVER w/o images & 66.3$ \pm $0.6 & 68.5$ \pm $0.6 & 61.6$ \pm $1.8 & 61.9$ \pm $0.7 & 45.4$ \pm $0.7 & 46.2$ \pm $0.4 & 67.5$ \pm $2.9 & 70.3$ \pm $1.1 \\
                \hline
                MEVER (ours) & \textbf{71.6$ \pm $0.7} & \textbf{71.6$ \pm $0.4} & \textbf{64.3$ \pm $0.6} & \textbf{64.1$ \pm $0.3} & \textbf{48.3$ \pm $2.1} & \textbf{49.7$ \pm $1.2} & \textbf{76.0$ \pm $0.7} & \textbf{77.7$ \pm $0.1} \\
			\bottomrule
		\end{tabular}
	}
	\label{table:claim_verification_macro_f1}
\end{table*}

\begin{table*}[t]
	\centering
	\caption{Explanation generation with \emph{ROUGE-L}, \emph{METEOR}, and \emph{BLEU-2} scores (\%) with \emph{retrieved evidence setting}. See Appendix \ref{sec:additional_explanation_generation} for results on gold evidence setting. MR2 dataset does not have explanations.}
	\resizebox{\textwidth}{!}{
		\begin{tabular}{c|ccc|ccc|ccc}
			\toprule
                \multirow{2}{*}{Model} & \multicolumn{3}{c|}{AIChartClaim} & \multicolumn{3}{c|}{ChartCheck} & \multicolumn{3}{c}{Mocheg} \\
			\cline{2-10}
			{} & ROUGE-L & METEOR & BLEU-2 & ROUGE-L & METEOR & BLEU-2 & ROUGE-L & METEOR & BLEU-2 \\
			\hline
                JustiLM & 21.7$ \pm $0.7 & 16.8$ \pm $1.0 & 12.4$ \pm $1.0 & 34.7$ \pm $1.8 & 30.6$ \pm $1.9 & 23.5$ \pm $1.9 & 19.3$ \pm $0.5 & 17.7$ \pm $0.3 & 12.1$ \pm $0.7 \\
                \hline
                MochegModel & 33.4$ \pm $1.0 & 25.7$ \pm $1.4 & 18.2$ \pm $1.5 & 39.6$ \pm $0.3 & 33.8$ \pm $0.4 & 24.5$ \pm $0.5 & 18.7$ \pm $0.4 & 18.7$ \pm $0.3 & 14.8$ \pm $0.1 \\
                ECENet & 32.3$ \pm $0.7 & 26.8$ \pm $0.6 & 19.2$ \pm $1.1 & 39.7$ \pm $0.6 & 34.7$ \pm $0.5 & 25.7$ \pm $0.8 & 20.5$ \pm $0.6 & 15.4$ \pm $0.1 & 12.2$ \pm $0.2 \\
                DePlot+FlanT5 & 33.5$ \pm $1.1 & 25.8$ \pm $1.4 & 18.3$ \pm $1.5 & 39.9$ \pm $0.1 & 34.0$ \pm $0.2 & 24.7$ \pm $0.2 & 19.7$ \pm $0.5 & 19.0$ \pm $0.5 & 16.2$ \pm $0.7 \\
                GPT-4o & 18.2$ \pm $0.6 & 23.2$ \pm $0.5 & 16.1$ \pm $0.3 & 17.7$ \pm $0.4 & 27.4$ \pm $0.2 & 16.2$ \pm $0.3 & 18.2$ \pm $1.6 & 18.8$ \pm $1.8 & 10.9$ \pm $0.8 \\
                \hline
                UniChart & 33.2$ \pm $0.2 & 25.9$ \pm $0.4 & 19.6$ \pm $0.1 & 40.1$ \pm $0.2 & 34.6$ \pm $0.2 & 26.6$ \pm $0.2 & N.A. & N.A. & N.A. \\
                ChartGemma & 33.5$ \pm $0.3 & 27.1$ \pm $0.2 & 20.4$ \pm $0.1 & 40.1$ \pm $0.1 & 35.7$ \pm $0.2 & 26.4$ \pm $0.1 & N.A. & N.A. & N.A. \\
                \hline
                MEVER w/o images & 33.7$ \pm $0.1 & 26.3$ \pm $0.6 & 21.1$ \pm $0.1 & 39.7$ \pm $0.1 & 34.8$ \pm $0.3 & 25.5$ \pm $0.5 & 22.9$ \pm $0.1 & 19.0$ \pm $0.1 & 15.1$ \pm $0.1 \\
                \hline
                MEVER (ours) & \textbf{34.5$ \pm $0.2} & \textbf{27.8$ \pm $0.4} & \textbf{21.3$ \pm $0.4 }& \textbf{40.8$ \pm $0.1} & \textbf{36.2$ \pm $0.2} & \textbf{27.2$ \pm $0.4} & \textbf{23.4$ \pm $0.3} & \textbf{20.0$ \pm $0.3} & \textbf{16.3$ \pm $0.3} \\
			\bottomrule
		\end{tabular}
	}
	\label{table:explanation_generation1}
\end{table*}

\begin{figure*}[t]
	\centering
	\includegraphics[width=1\linewidth]{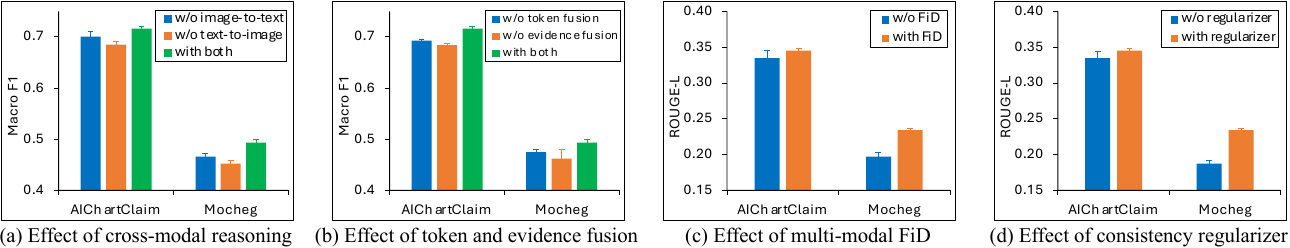}
	\caption{Model analysis on AIChartClaim and Mocheg datasets.}
	\label{fig:model_analysis}
\end{figure*}

\begin{figure}[t]
	\centering
	\includegraphics[width=1\linewidth]{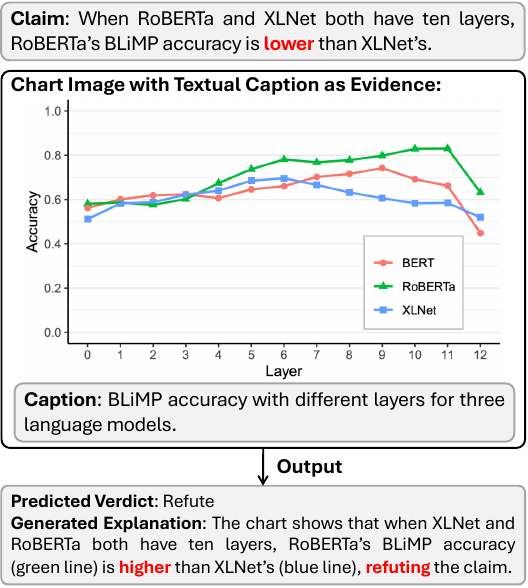}
	\caption{Case study on AIChartClaim dataset.}
	\label{fig:case_study}
\end{figure}

\textbf{Datasets}. We present main statistics of the datasets in Table \ref{table:dataset_statistics}. Besides \textbf{AIChartClaim}, we have \textbf{ChartCheck} \cite{chartcheck}, a general-domain chart dataset. \textbf{Mocheg} \cite{mocheg} is multi-modal dataset with explanation. \textbf{MR2} \cite{mr2} has images for claims and evidence, but no explanations. Appendix \ref{sec:additional_dataset_details} has more details.

\textbf{Implementation.} 
We set $ L $ to 12 and $ d $ to 768. 
We initialize the model with scientific parameters \cite{scibert} for AIChartClaim and with general parameters \cite{bert} for others. Each result is obtained by 3 independent runs. Experiments are done on 4 NVIDIA A100 GPUs. More details are in Appendix \ref{sec:experiment_environment}.

\subsection{Evidence Retrieval}


\textbf{Baselines.} \textbf{\emph{i}) Text-only retrieval}, 
BM25 \cite{bm25}, RAV \cite{rav}, JustiLM \cite{justilm}. 
\textbf{\emph{ii}) Multi-modal retrieval}, 
MochegModel \cite{mocheg}. Since our retriever is built on TransformerXH and ViT, we add one multi-modal baseline, TransXH+ViT. 
We add an ablation by removing images from our model.

\textbf{Analysis.} If the dataset has training-test split, we follow its split. Otherwise, we split 80\% for training, among which 10\% for validation. We follow \citet{mocheg} and report MAP, Precision@$ \kappa $, and Recall@$ \kappa $ in Table \ref{table:evidence_retrieval} with $ \kappa=3 $. We further vary $ \kappa $ in \{1, 3, 5, 7\} in Appendix \ref{sec:additional_evidence_retrieval}. Multi-modal retrieval outperforms text-only methods, since images bring useful information. Our model performs better than them, since our multi-modal graph encoder uses a nested architecture to integrate multi-modal data. The ablated model drops the result, verifying that images provide useful information.

\subsection{Claim Verification}
\label{sec:claim_verification}

\textbf{Baselines.} \textbf{\emph{i}) Text-only}, KGAT \cite{kgat}, CausalWalk \cite{causalwalk}, CORRECT \cite{correct}, TransformerXH \cite{transformer_xh}. \textbf{\emph{ii}) Multi-modal}, MochegModel \cite{mocheg}, MR2Retrieved \cite{mr2}, CCN \cite{ccn}, ESCNet \cite{escnet}, ECENet \cite{ecenet}, MultiKE-GAT \cite{multike_gat}, GPT-4o \cite{gpt4}. \textbf{\emph{iii}) Chart-based}, ChartBERT \cite{chartbert}, UniChart \cite{unichart}, ChartGemma \cite{chartgemma}. \textbf{\emph{iv}) Retrieval-augmented}, 
JustiLM \cite{justilm}, RAV \cite{rav}. We convert TransformerXH to its multi-modal version, TransformerXH++. We add our ablated model by removing images. For instruction-tuned models, we use 5-shot setting. Since chart-based models are specifically designed for charts, we do not report their result on Mocheg and MR2.

\textbf{Gold v.s. retrieved setting.} For gold setting, we have gold multi-modal evidence. For retrieved setting, we first retrieve multi-modal evidence, which is then used for verification. Since our retriever achieves the best result, we use our retrieved evidence for our model and all baselines for fairness.

\textbf{Analysis.} As in \citet{multivers}, we show Macro F1 in Table \ref{table:claim_verification_macro_f1}. See Appendix \ref{sec:additional_claim_verification} for Micro F1. Multi-modal baselines tend to outperform others, showing the strength of images. However, all of them model evidence-level fusion. Our model is better than them, showing the strength of both token- and evidence-level fusion. 
Though evidence retrieval on Mocheg and MR2 is more difficult, MEVER still outperforms baselines on retrieved setting, showing that MEVER is not affected by the difficulty of evidence retrieval. GPT-4o does not perform well on AIChartClaim, since our dataset creation inputs gold label to GPT-4o to generate explanations, but the current verification does not input gold labels, leading to inaccurate result.

See Table \ref{table:ai_chart_check_results} for more results on AIChartClaim.

\subsection{Explanation Generation}

\textbf{Baselines.} \textbf{\emph{i}) Text-only}, JustiLM \cite{justilm}, a retrieval-augmented baseline. 
\textbf{\emph{ii}) Multi-modal}, MochegModel \cite{mocheg}, ECENet \cite{ecenet}, DePlot+FlanT5 \cite{chartcheck}, GPT-4o \cite{gpt4}. \textbf{\emph{iii}) Chart-based}, UniChart \cite{unichart}, ChartGemma \cite{chartgemma}. 
We add our ablated model.

\textbf{Analysis.} We follow \citet{mocheg} and show ROUGE-L, METEOR, and BLEU-2 with retrieved setting in Table \ref{table:explanation_generation1}. See Appendix \ref{sec:additional_explanation_generation} for ROUGE-1, ROUGE-2, BLEU-4, and VLM-as-a-Judge. 
Our model outperforms baselines, since consistency regularizer uses predicted label to regularize the explanation towards accurate generation. GPT-4o does not perform well, because we use its predicted labels in Sec. \ref{sec:claim_verification} as input to generate explanation to keep consistent with our model. Its inaccurate predicted verdicts influence explanation generation.

\subsection{Model Analysis}

\textbf{Cross-modal reasoning.} We respectively remove image-to-text and text-to-image reasoning from the model in Fig. \ref{fig:model_analysis}(a). The complete model performs the best, showing the strength of both reasoning. Removing text-to-image leads to the worst result, since images do not contain enough information. 

\textbf{Token and evidence fusion.} We remove each fusion in Fig. \ref{fig:model_analysis}(b). The complete model outperforms others, showing the strength of both fusions. Removing evidence fusion hurts the result, since claim embedding, though with evidence information after claim-evidence interaction, still needs evidence explicitly for accurate verification.

\textbf{Multi-modal Fusion-in-Decoder (FiD).} Fig. \ref{fig:model_analysis}(c) shows that removing FiD hurts explanation generation quality, since FiD aggregates multiple evidence for generation, and disregarding it leads to insufficient multi-evidence reasoning.

\textbf{Consistency regularizer.} Fig. \ref{fig:model_analysis}(d) shows that consistency regularizer controls the generation towards the predicted label for consistent explanation, thus improving explanation quality.

\textbf{Case study.} In Fig. \ref{fig:case_study}, our model reasons over text and image to correctly verify the claim, and the explanation clearly justifies the prediction. This visualization shows the effectiveness of our model. For more case studies, our submitted code produces predicted labels and explanations for all claims.

\subsection{Analysis of the AIChartClaim Dataset}


\textbf{Chart type.} We collect 300 charts in total and provide chart types in Table \ref{table:ai_chart_claim_dataset_statistics}. Since line charts and bar charts are the most commonly used charts in academic papers to express scientific results and discoveries, they are the top-2 chart types in the dataset. Here ``Bar with Number'' indicates a bar chart where authors also put numerical result on top of each bar, same for ``Line with Number''. We differentiate them from bar and line charts, because usually models need to further recognize the numerical value on the bar and line to verify the claims.

\textbf{Analysis on each chart type.} We further report claim verification result of our model on each chart type in Table \ref{table:ai_chart_check_results}. Overall, our model performs the best on line charts, including ``Line'' and ``Line with Number''. One reason is that our dataset has the most number of line charts, which contain more information than other chart types to train the model. The performance on bar charts, including ``Bar'', ``Bar with Number'', and ``Line + Bar'', is also decent. Our model does not perform well on other chart types, 
potentially due to insufficient training data. However, we emphasize that line charts and bar charts are the most common charts in AI papers, and it is quite difficult to obtain sufficient scatterplots, pie charts, and heat maps from AI papers. 

\begin{table}[t]
	\centering
	\caption{AIChartClaim dataset statistics.}
	\resizebox{0.9\columnwidth}{!}{
		\begin{tabular}{c|c}
			\toprule
			Chart Type & Number of Charts \\
                \hline
                Line & 203 \\
                Bar & 61 \\
                Bar with Number & 16 \\
                Line with Number & 6 \\
                Line + Bar & 5 \\
                Scatterplot & 4 \\
                Scatterplot with Lines & 3 \\
                Others (Pie and Heat Map) & 2 \\
                \hline
                Total & 300 \\
			\bottomrule
		\end{tabular}
	}
	\label{table:ai_chart_claim_dataset_statistics}
\end{table}

\begin{table}[t]
	\centering
	\caption{Claim verification of our model MEVER on test set of each chart type (\%).}
	\resizebox{\columnwidth}{!}{
		\begin{tabular}{c|cc}
			\toprule
                \multirow{2}{*}{Chart Type} & \multicolumn{2}{c}{AIChartClaim} \\
			\cline{2-3}
			{} & Micro F1 & Macro F1 \\
			\hline
                Line & 75.6 & 75.6 \\
                Bar & 67.3 & 67.0 \\
                Bar with Number & 56.3 & 56.3 \\
                Line with Number & 75.0 & 75.3 \\
                Line + Bar & 62.5 & 61.9 \\
                Others (Scatterplot, Pie, Heat Map) & 55.6 & 55.6 \\
			\bottomrule
		\end{tabular}
	}
	\label{table:ai_chart_check_results}
\end{table}
\section{Conclusion}

We propose MEVER with evidence retrieval, claim verification, and explanation. We create a scientific dataset in AI domain. A future work is to explore multi-modal knowledge graph for verification.
\section*{Acknowledgments}

This work was in part supported by NSF awards \#1934782 and \#2114824. Some of the research results were obtained using computational resources provided by NAIRR award \#240336.

\clearpage
\section*{Limitations}

Here we identify two limitations in terms of evidence type and our proposed AIChartClaim dataset.

\textbf{Evidence type.} Our model is proposed mainly in the multi-modal setting where we assume both texts and images are available. If images are absent from the dataset and we have texts only, we may need additional effort to obtain images, so that our model can use both modalities for claim verification. As suggested by \citet{ccn}, one potential solution is to search for relevant images given evidence texts, such as collecting charts using the captions or paper titles. Since our paper does not study such cross-modal information search, we leave it as a future work.

\textbf{AIChartClaim dataset.} \textbf{\emph{i}) Size.} The size of our created dataset is already comparable to other scientific text-only datasets. The size of our dataset is limited by two main factors. First, the number of AI papers with checkworthy claims and clearly readable charts is limited. We have tried our best to obtain appropriate AI papers. Second, the dataset creation process requires domain experts with scientific knowledge to check and analyze the charts, thus the creation is quite effort-consuming and knowledge-dependent. The number of domain experts is also limited. We consider creating a large-size dataset as a future work.

\textbf{\emph{ii}) Domain.} We are AI researchers and mainly focus on AI domain for now. Since there is not an existing scientific dataset with multi-modal data and explanation, our dataset is the first step to advance the research community. We are interested in collecting datasets in other domains, such as biomedicine, in the future.

\textbf{\emph{iii}) Verdict.} Following existing chart datasets, e.g., ChartCheck \cite{chartcheck}, our dataset has two types of verdicts, SUPPORT and REFUTE. Some other datasets have the third verdict, NEI for Not Enough Info. Following \citet{check_covid}, we can also introduce the third verdict by simply replacing the gold evidence of claims with random evidence. But for clarity and consistency purpose, we do not introduce NEI verdict in our dataset.
\section*{Ethics Statement}

We do not foresee any undesired implications stemming from our work. Conversely, we hope that our work can advance AI Ethics research.

\bibliography{latex/acl_latex}

\begin{thebibliography}{68}
\providecommand{\natexlab}[1]{#1}

\bibitem[{Abdelnabi et~al.(2022)Abdelnabi, Hasan, and Fritz}]{ccn}
Sahar Abdelnabi, Rakibul Hasan, and Mario Fritz. 2022.
\newblock Open-domain, content-based, multi-modal fact-checking of out-of-context images via online resources.
\newblock In \emph{Proceedings of the IEEE/CVF conference on computer vision and pattern recognition}, pages 14940--14949.

\bibitem[{Achiam et~al.(2023)Achiam, Adler, Agarwal, Ahmad, Akkaya, Aleman, Almeida, Altenschmidt, Altman, Anadkat et~al.}]{gpt4}
Josh Achiam, Steven Adler, Sandhini Agarwal, Lama Ahmad, Ilge Akkaya, Florencia~Leoni Aleman, Diogo Almeida, Janko Altenschmidt, Sam Altman, Shyamal Anadkat, et~al. 2023.
\newblock Gpt-4 technical report.
\newblock \emph{arXiv preprint arXiv:2303.08774}.

\bibitem[{Akhtar et~al.(2023)Akhtar, Cocarascu, and Simperl}]{chartbert}
Mubashara Akhtar, Oana Cocarascu, and Elena Simperl. 2023.
\newblock Reading and reasoning over chart images for evidence-based automated fact-checking.
\newblock In \emph{Findings of the Association for Computational Linguistics: EACL 2023}, pages 399--414.

\bibitem[{Akhtar et~al.(2024)Akhtar, Subedi, Gupta, Tahmasebi, Cocarascu, and Simperl}]{chartcheck}
Mubashara Akhtar, Nikesh Subedi, Vivek Gupta, Sahar Tahmasebi, Oana Cocarascu, and Elena Simperl. 2024.
\newblock Chartcheck: Explainable fact-checking over real-world chart images.
\newblock In \emph{Findings of the Association for Computational Linguistics ACL 2024}, pages 13921--13937.

\bibitem[{Aly et~al.(2021)Aly, Guo, Schlichtkrull, Thorne, Vlachos, Christodoulopoulos, Cocarascu, and Mittal}]{feverous}
Rami Aly, Zhijiang Guo, Michael Schlichtkrull, James Thorne, Andreas Vlachos, Christos Christodoulopoulos, Oana Cocarascu, and Arpit Mittal. 2021.
\newblock Feverous: Fact extraction and verification over unstructured and structured information.
\newblock In \emph{35th Conference on Neural Information Processing Systems, NeurIPS 2021}. Neural Information Processing Systems foundation.

\bibitem[{Atanasova(2024)}]{explainable1}
Pepa Atanasova. 2024.
\newblock Generating fact checking explanations.
\newblock In \emph{Accountable and Explainable Methods for Complex Reasoning over Text}, pages 83--103. Springer.

\bibitem[{Beltagy et~al.(2019)Beltagy, Lo, and Cohan}]{scibert}
Iz~Beltagy, Kyle Lo, and Arman Cohan. 2019.
\newblock Scibert: A pretrained language model for scientific text.
\newblock In \emph{Proceedings of the 2019 Conference on Empirical Methods in Natural Language Processing and the 9th International Joint Conference on Natural Language Processing (EMNLP-IJCNLP)}, pages 3615--3620.

\bibitem[{Beltagy et~al.(2020)Beltagy, Peters, and Cohan}]{longformer}
Iz~Beltagy, Matthew~E Peters, and Arman Cohan. 2020.
\newblock Longformer: The long-document transformer.
\newblock \emph{arXiv preprint arXiv:2004.05150}.

\bibitem[{Bishop and Nasrabadi(2006)}]{prml}
Christopher~M Bishop and Nasser~M Nasrabadi. 2006.
\newblock \emph{Pattern recognition and machine learning}, volume~4.
\newblock Springer.

\bibitem[{Cao et~al.(2024)Cao, Wei, Zhou, and Hu}]{multike_gat}
Han Cao, Lingwei Wei, Wei Zhou, and Songlin Hu. 2024.
\newblock Multi-source knowledge enhanced graph attention networks for multimodal fact verification.
\newblock In \emph{2024 IEEE International Conference on Multimedia and Expo (ICME)}, pages 1--6. IEEE.

\bibitem[{Devlin et~al.(2019)Devlin, Chang, Lee, and Toutanova}]{bert}
Jacob Devlin, Ming-Wei Chang, Kenton Lee, and Kristina Toutanova. 2019.
\newblock Bert: Pre-training of deep bidirectional transformers for language understanding.
\newblock In \emph{Proceedings of the 2019 Conference of the North American Chapter of the Association for Computational Linguistics: Human Language Technologies, Volume 1 (Long and Short Papers)}, pages 4171--4186.

\bibitem[{Dosovitskiy et~al.(2021)Dosovitskiy, Beyer, Kolesnikov, Weissenborn, Zhai, Unterthiner, Dehghani, Minderer, Heigold, Gelly, Uszkoreit, and Houlsby}]{vit}
Alexey Dosovitskiy, Lucas Beyer, Alexander Kolesnikov, Dirk Weissenborn, Xiaohua Zhai, Thomas Unterthiner, Mostafa Dehghani, Matthias Minderer, Georg Heigold, Sylvain Gelly, Jakob Uszkoreit, and Neil Houlsby. 2021.
\newblock An image is worth 16x16 words: Transformers for image recognition at scale.
\newblock In \emph{International Conference on Learning Representations}.

\bibitem[{Hamilton et~al.(2017)Hamilton, Ying, and Leskovec}]{graphsage}
Will Hamilton, Zhitao Ying, and Jure Leskovec. 2017.
\newblock Inductive representation learning on large graphs.
\newblock \emph{Advances in neural information processing systems}, 30.

\bibitem[{Hu et~al.(2023)Hu, Guo, Chen, Wen, and Yu}]{mr2}
Xuming Hu, Zhijiang Guo, Junzhe Chen, Lijie Wen, and Philip~S Yu. 2023.
\newblock Mr2: A benchmark for multimodal retrieval-augmented rumor detection in social media.
\newblock In \emph{Proceedings of the 46th international ACM SIGIR conference on research and development in information retrieval}, pages 2901--2912.

\bibitem[{Huang et~al.(2024)Huang, Chan, Fung, Qiu, Zhou, Joty, Chang, and Ji}]{chart_survey}
Kung-Hsiang Huang, Hou~Pong Chan, Yi~R Fung, Haoyi Qiu, Mingyang Zhou, Shafiq Joty, Shih-Fu Chang, and Heng Ji. 2024.
\newblock From pixels to insights: A survey on automatic chart understanding in the era of large foundation models.
\newblock \emph{IEEE Transactions on Knowledge and Data Engineering}.

\bibitem[{Izacard et~al.(2023)Izacard, Lewis, Lomeli, Hosseini, Petroni, Schick, Dwivedi-Yu, Joulin, Riedel, and Grave}]{atlas}
Gautier Izacard, Patrick Lewis, Maria Lomeli, Lucas Hosseini, Fabio Petroni, Timo Schick, Jane Dwivedi-Yu, Armand Joulin, Sebastian Riedel, and Edouard Grave. 2023.
\newblock Atlas: Few-shot learning with retrieval augmented language models.
\newblock \emph{Journal of Machine Learning Research}, 24(251):1--43.

\bibitem[{Kotonya and Toni(2020{\natexlab{a}})}]{explainable2}
Neema Kotonya and Francesca Toni. 2020{\natexlab{a}}.
\newblock Explainable automated fact-checking for public health claims.
\newblock In \emph{Proceedings of the 2020 Conference on Empirical Methods in Natural Language Processing (EMNLP)}, pages 7740--7754.

\bibitem[{Kotonya and Toni(2020{\natexlab{b}})}]{pubhealth}
Neema Kotonya and Francesca Toni. 2020{\natexlab{b}}.
\newblock Explainable automated fact-checking for public health claims.
\newblock In \emph{Proceedings of the 2020 Conference on Empirical Methods in Natural Language Processing (EMNLP)}, pages 7740--7754.

\bibitem[{Lee et~al.(2024)Lee, Kim, Park, Kim, and Seo}]{vlm_as_a_judge}
Seongyun Lee, Seungone Kim, Sue Park, Geewook Kim, and Minjoon Seo. 2024.
\newblock Prometheus-vision: Vision-language model as a judge for fine-grained evaluation.
\newblock In \emph{Findings of the Association for Computational Linguistics ACL 2024}, pages 11286--11315.

\bibitem[{Liu et~al.(2023{\natexlab{a}})Liu, Piccinno, Krichene, Pang, Lee, Joshi, Altun, Collier, and Eisenschlos}]{matcha}
Fangyu Liu, Francesco Piccinno, Syrine Krichene, Chenxi Pang, Kenton Lee, Mandar Joshi, Yasemin Altun, Nigel Collier, and Julian~Martin Eisenschlos. 2023{\natexlab{a}}.
\newblock Matcha: Enhancing visual language pretraining with math reasoning and chart derendering.
\newblock In \emph{The 61st Annual Meeting Of The Association For Computational Linguistics}.

\bibitem[{Liu et~al.(2024{\natexlab{a}})Liu, Wang, Yao, Chen, Song, Cho, Yacoob, and Yu}]{mmc}
Fuxiao Liu, Xiaoyang Wang, Wenlin Yao, Jianshu Chen, Kaiqiang Song, Sangwoo Cho, Yaser Yacoob, and Dong Yu. 2024{\natexlab{a}}.
\newblock Mmc: Advancing multimodal chart understanding with large-scale instruction tuning.
\newblock In \emph{Proceedings of the 2024 Conference of the North American Chapter of the Association for Computational Linguistics: Human Language Technologies (Volume 1: Long Papers)}, pages 1287--1310.

\bibitem[{Liu et~al.(2024{\natexlab{b}})Liu, Li, Li, and Lee}]{llava2}
Haotian Liu, Chunyuan Li, Yuheng Li, and Yong~Jae Lee. 2024{\natexlab{b}}.
\newblock Improved baselines with visual instruction tuning.
\newblock In \emph{Proceedings of the IEEE/CVF Conference on Computer Vision and Pattern Recognition}, pages 26296--26306.

\bibitem[{Liu et~al.(2023{\natexlab{b}})Liu, Li, Wu, and Lee}]{llava}
Haotian Liu, Chunyuan Li, Qingyang Wu, and Yong~Jae Lee. 2023{\natexlab{b}}.
\newblock Visual instruction tuning.
\newblock \emph{Advances in neural information processing systems}, 36:34892--34916.

\bibitem[{Liu et~al.(2020)Liu, Xiong, Sun, and Liu}]{kgat}
Zhenghao Liu, Chenyan Xiong, Maosong Sun, and Zhiyuan Liu. 2020.
\newblock Fine-grained fact verification with kernel graph attention network.
\newblock In \emph{Proceedings of the 58th Annual Meeting of the Association for Computational Linguistics}, pages 7342--7351.

\bibitem[{Luo et~al.(2021)Luo, Darrell, and Rohrbach}]{newsclippings}
Grace Luo, Trevor Darrell, and Anna Rohrbach. 2021.
\newblock Newsclippings: Automatic generation of out-of-context multimodal media.
\newblock In \emph{Proceedings of the 2021 Conference on Empirical Methods in Natural Language Processing}, pages 6801--6817.

\bibitem[{Luo et~al.(2025)Luo, Chen, Zheng, Wu, Guo, Lin, Feng, Kuang, Song, Zhu et~al.}]{hypergraphrag}
Haoran Luo, Guanting Chen, Yandan Zheng, Xiaobao Wu, Yikai Guo, Qika Lin, Yu~Feng, Zemin Kuang, Meina Song, Yifan Zhu, et~al. 2025.
\newblock Hypergraphrag: Retrieval-augmented generation via hypergraph-structured knowledge representation.
\newblock \emph{arXiv preprint arXiv:2503.21322}.

\bibitem[{Masry et~al.(2023)Masry, Kavehzadeh, Do, Hoque, and Joty}]{unichart}
Ahmed Masry, Parsa Kavehzadeh, Xuan~Long Do, Enamul Hoque, and Shafiq Joty. 2023.
\newblock Unichart: A universal vision-language pretrained model for chart comprehension and reasoning.
\newblock In \emph{Proceedings of the 2023 Conference on Empirical Methods in Natural Language Processing}, pages 14662--14684.

\bibitem[{Masry et~al.(2025)Masry, Thakkar, Bajaj, Kartha, Hoque, and Joty}]{chartgemma}
Ahmed Masry, Megh Thakkar, Aayush Bajaj, Aaryaman Kartha, Enamul Hoque, and Shafiq Joty. 2025.
\newblock Chartgemma: Visual instruction-tuning for chart reasoning in the wild.
\newblock In \emph{Proceedings of the 31st International Conference on Computational Linguistics: Industry Track}, pages 625--643.

\bibitem[{Meng et~al.(2024)Meng, Shao, Lu, Gao, Zhang, Qiao, and Luo}]{chartassistant}
Fanqing Meng, Wenqi Shao, Quanfeng Lu, Peng Gao, Kaipeng Zhang, Yu~Qiao, and Ping Luo. 2024.
\newblock Chartassistant: A universal chart multimodal language model via chart-to-table pre-training and multitask instruction tuning.
\newblock In \emph{Findings of the Association for Computational Linguistics ACL 2024}, pages 7775--7803.

\bibitem[{Mishra et~al.(2022)Mishra, Suryavardan, Bhaskar, Chopra, Reganti, Patwa, Das, Chakraborty, Sheth, Ekbal et~al.}]{factify}
Shreyash Mishra, S~Suryavardan, Amrit Bhaskar, Parul Chopra, Aishwarya~N Reganti, Parth Patwa, Amitava Das, Tanmoy Chakraborty, Amit~P Sheth, Asif Ekbal, et~al. 2022.
\newblock Factify: A multi-modal fact verification dataset.
\newblock In \emph{DE-FACTIFY@ AAAI}.

\bibitem[{Nguyen et~al.(2025{\natexlab{a}})Nguyen, Wu, Vu, Zhao, Nguyen, and Luu}]{cutpaste}
Cong-Duy Nguyen, Xiaobao Wu, Duc~Anh Vu, Shuai Zhao, Thong Nguyen, and Anh~Tuan Luu. 2025{\natexlab{a}}.
\newblock Cutpaste\&find: Efficient multimodal hallucination detector with visual-aid knowledge base.
\newblock \emph{arXiv preprint arXiv:2502.12591}.

\bibitem[{Nguyen et~al.(2025{\natexlab{b}})Nguyen, Wu, Bin, Nguyen, Ng, and Luu}]{scene_graph}
Thong~Thanh Nguyen, Xiaobao Wu, Yi~Bin, Cong-Duy~T Nguyen, See-Kiong Ng, and Anh~Tuan Luu. 2025{\natexlab{b}}.
\newblock Motion-aware contrastive learning for temporal panoptic scene graph generation.
\newblock In \emph{Proceedings of the AAAI Conference on Artificial Intelligence}, volume~39, pages 6218--6226.

\bibitem[{Pan et~al.(2023)Pan, Wu, Lu, Tuan, Wang, Kan, and Nakov}]{program_fc}
Liangming Pan, Xiaobao Wu, Xinyuan Lu, Luu~Anh Tuan, William~Yang Wang, Min-Yen Kan, and Preslav Nakov. 2023.
\newblock Fact-checking complex claims with program-guided reasoning.
\newblock In \emph{Proceedings of the 61st annual meeting of the association for computational linguistics (volume 1: long papers)}, pages 6981--7004.

\bibitem[{Radford et~al.(2021)Radford, Kim, Hallacy, Ramesh, Goh, Agarwal, Sastry, Askell, Mishkin, Clark et~al.}]{clip}
Alec Radford, Jong~Wook Kim, Chris Hallacy, Aditya Ramesh, Gabriel Goh, Sandhini Agarwal, Girish Sastry, Amanda Askell, Pamela Mishkin, Jack Clark, et~al. 2021.
\newblock Learning transferable visual models from natural language supervision.
\newblock In \emph{International conference on machine learning}, pages 8748--8763. PMLR.

\bibitem[{Raffel et~al.(2020)Raffel, Shazeer, Roberts, Lee, Narang, Matena, Zhou, Li, and Liu}]{t5}
Colin Raffel, Noam Shazeer, Adam Roberts, Katherine Lee, Sharan Narang, Michael Matena, Yanqi Zhou, Wei Li, and Peter~J Liu. 2020.
\newblock Exploring the limits of transfer learning with a unified text-to-text transformer.
\newblock \emph{Journal of machine learning research}, 21(140):1--67.

\bibitem[{Robertson et~al.(2009)Robertson, Zaragoza et~al.}]{bm25}
Stephen Robertson, Hugo Zaragoza, et~al. 2009.
\newblock The probabilistic relevance framework: Bm25 and beyond.
\newblock \emph{Foundations and Trends{\textregistered} in Information Retrieval}, 3(4):333--389.

\bibitem[{Russo et~al.(2023)Russo, Tekiro{\u{g}}lu, and Guerini}]{explainable4}
Daniel Russo, Serra~Sinem Tekiro{\u{g}}lu, and Marco Guerini. 2023.
\newblock Benchmarking the generation of fact checking explanations.
\newblock \emph{Transactions of the Association for Computational Linguistics}, 11:1250--1264.

\bibitem[{Schlichtkrull et~al.(2024)Schlichtkrull, Guo, and Vlachos}]{averitec}
Michael Schlichtkrull, Zhijiang Guo, and Andreas Vlachos. 2024.
\newblock Averitec: A dataset for real-world claim verification with evidence from the web.
\newblock \emph{Advances in Neural Information Processing Systems}, 36.

\bibitem[{Subramanian and Lee(2020)}]{hesm}
Shyam Subramanian and Kyumin Lee. 2020.
\newblock Hierarchical evidence set modeling for automated fact extraction and verification.
\newblock In \emph{Proceedings of the 2020 Conference on Empirical Methods in Natural Language Processing (EMNLP)}, pages 7798--7809.

\bibitem[{Thorne et~al.(2018)Thorne, Vlachos, Christodoulopoulos, and Mittal}]{fever}
J~Thorne, A~Vlachos, C~Christodoulopoulos, and A~Mittal. 2018.
\newblock Fever: a large-scale dataset for fact extraction and verification.
\newblock In \emph{Proceedings of the 16th Annual Conference of the North American Chapter of the Association for Computational Linguistics: Human Language Technologies}. Sheffield.

\bibitem[{Vaswani et~al.(2017)Vaswani, Shazeer, Parmar, Uszkoreit, Jones, Gomez, Kaiser, and Polosukhin}]{transformer}
Ashish Vaswani, Noam Shazeer, Niki Parmar, Jakob Uszkoreit, Llion Jones, Aidan~N Gomez, {\L}ukasz Kaiser, and Illia Polosukhin. 2017.
\newblock Attention is all you need.
\newblock \emph{Advances in neural information processing systems}, 30.

\bibitem[{Veli{\v{c}}kovi{\'c} et~al.(2018)Veli{\v{c}}kovi{\'c}, Cucurull, Casanova, Romero, Li{\`o}, and Bengio}]{gat}
Petar Veli{\v{c}}kovi{\'c}, Guillem Cucurull, Arantxa Casanova, Adriana Romero, Pietro Li{\`o}, and Yoshua Bengio. 2018.
\newblock Graph attention networks.
\newblock In \emph{International Conference on Learning Representations}.

\bibitem[{Wadden et~al.(2020)Wadden, Lin, Lo, Wang, van Zuylen, Cohan, and Hajishirzi}]{scifact}
David Wadden, Shanchuan Lin, Kyle Lo, Lucy~Lu Wang, Madeleine van Zuylen, Arman Cohan, and Hannaneh Hajishirzi. 2020.
\newblock Fact or fiction: Verifying scientific claims.
\newblock In \emph{Proceedings of the 2020 Conference on Empirical Methods in Natural Language Processing (EMNLP)}, pages 7534--7550.

\bibitem[{Wadden et~al.(2022)Wadden, Lo, Wang, Cohan, Beltagy, and Hajishirzi}]{multivers}
David Wadden, Kyle Lo, Lucy~Lu Wang, Arman Cohan, Iz~Beltagy, and Hannaneh Hajishirzi. 2022.
\newblock Multivers: Improving scientific claim verification with weak supervision and full-document context.
\newblock In \emph{Findings of the Association for Computational Linguistics: NAACL 2022}, pages 61--76.

\bibitem[{Wang et~al.(2023)Wang, Harwood, Chillrud, Ananthram, Subbiah, and Mckeown}]{check_covid}
Gengyu Wang, Kate Harwood, Lawrence Chillrud, Amith Ananthram, Melanie Subbiah, and Kathleen Mckeown. 2023.
\newblock Check-covid: Fact-checking covid-19 news claims with scientific evidence.
\newblock In \emph{Findings of the Association for Computational Linguistics: ACL 2023}, pages 14114--14127.

\bibitem[{Wang et~al.(2025)Wang, Rangapur, Xu, Liang, Gharwi, Yang, and Shu}]{mmcv}
Haoran Wang, Aman Rangapur, Xiongxiao Xu, Yueqing Liang, Haroon Gharwi, Carl Yang, and Kai Shu. 2025.
\newblock Piecing it all together: Verifying multi-hop multimodal claims.
\newblock In \emph{Proceedings of the 31st International Conference on Computational Linguistics}, pages 7453--7469.

\bibitem[{Wu et~al.(2024{\natexlab{a}})Wu, Pan, Wang, and Luu}]{uke}
Xiaobao Wu, Liangming Pan, William~Yang Wang, and Anh~Tuan Luu. 2024{\natexlab{a}}.
\newblock Updating language models with unstructured facts: Towards practical knowledge editing.
\newblock \emph{CoRR}.

\bibitem[{Wu et~al.(2024{\natexlab{b}})Wu, Pan, Wang, and Tuan}]{akew}
Xiaobao Wu, Liangming Pan, William~Yang Wang, and Luu~Anh Tuan. 2024{\natexlab{b}}.
\newblock Akew: Assessing knowledge editing in the wild.
\newblock In \emph{Proceedings of the 2024 Conference on Empirical Methods in Natural Language Processing}, pages 15118--15133.

\bibitem[{Wuehrl et~al.(2024)Wuehrl, Resendiz, Grimminger, and Klinger}]{bear_fact}
Amelie Wuehrl, Yarik~Menchaca Resendiz, Lara Grimminger, and Roman Klinger. 2024.
\newblock What makes medical claims (un) verifiable? analyzing entity and relation properties for fact verification.
\newblock In \emph{Proceedings of the 18th Conference of the European Chapter of the Association for Computational Linguistics (Volume 1: Long Papers)}, pages 2046--2058.

\bibitem[{Yamashita et~al.(2025)Yamashita, Tran, Zhang, and Lee}]{fake_resume}
Michiharu Yamashita, Thanh Tran, Delvin~Ce Zhang, and Dongwon Lee. 2025.
\newblock Unmasking fake careers: Detecting machine-generated career trajectories via multi-layer heterogeneous graphs.
\newblock In \emph{Proceedings of the 2025 Conference on Empirical Methods in Natural Language Processing}, pages 20893--20908.

\bibitem[{Yao et~al.(2023)Yao, Shah, Sun, Cho, and Huang}]{mocheg}
Barry~Menglong Yao, Aditya Shah, Lichao Sun, Jin-Hee Cho, and Lifu Huang. 2023.
\newblock End-to-end multimodal fact-checking and explanation generation: A challenging dataset and models.
\newblock In \emph{Proceedings of the 46th International ACM SIGIR Conference on Research and Development in Information Retrieval}, pages 2733--2743.

\bibitem[{Yue et~al.(2024)Yue, Zeng, Shang, Liu, Zhang, and Wang}]{rafts}
Zhenrui Yue, Huimin Zeng, Lanyu Shang, Yifan Liu, Yang Zhang, and Dong Wang. 2024.
\newblock Retrieval augmented fact verification by synthesizing contrastive arguments.
\newblock In \emph{Proceedings of the 62nd Annual Meeting of the Association for Computational Linguistics (Volume 1: Long Papers)}, pages 10331--10343, Bangkok, Thailand. Association for Computational Linguistics.

\bibitem[{Zeng and Gao(2024)}]{justilm}
Fengzhu Zeng and Wei Gao. 2024.
\newblock Justilm: Few-shot justification generation for explainable fact-checking of real-world claims.
\newblock \emph{Transactions of the Association for Computational Linguistics}, 12:334--354.

\bibitem[{Zhang and Lauw(2020)}]{adjacent_encoder}
Ce~Zhang and Hady~W Lauw. 2020.
\newblock Topic modeling on document networks with adjacent-encoder.
\newblock In \emph{Proceedings of the AAAI Conference on Artificial Intelligence}, volume~34, pages 6737--6745.

\bibitem[{Zhang et~al.(2024{\natexlab{a}})Zhang, Zhang, and Zhou}]{causalwalk}
Congzhi Zhang, Linhai Zhang, and Deyu Zhou. 2024{\natexlab{a}}.
\newblock Causal walk: Debiasing multi-hop fact verification with front-door adjustment.
\newblock In \emph{Proceedings of the AAAI Conference on Artificial Intelligence}, volume~38, pages 19533--19541.

\bibitem[{Zhang and Lauw(2023)}]{dbn}
Delvin~Ce Zhang and Hady~W Lauw. 2023.
\newblock Topic modeling on document networks with dirichlet optimal transport barycenter.
\newblock \emph{IEEE Transactions on Knowledge and Data Engineering}, 36(3):1328--1340.

\bibitem[{Zhang and Lee(2025)}]{correct}
Delvin~Ce Zhang and Dongwon Lee. 2025.
\newblock Correct: Context- and reference-augmented reasoning and prompting for fact-checking.
\newblock In \emph{Proceedings of 2025 Annual Conference of the Nations of the Americas Chapter of the Association for Computational Linguistics}.

\bibitem[{Zhang et~al.(2023{\natexlab{a}})Zhang, Ying, and Lauw}]{hgtm}
Delvin~Ce Zhang, Rex Ying, and Hady~W Lauw. 2023{\natexlab{a}}.
\newblock Hyperbolic graph topic modeling network with continuously updated topic tree.
\newblock In \emph{Proceedings of the 29th ACM SIGKDD Conference on Knowledge Discovery and Data Mining}, pages 3206--3216.

\bibitem[{Zhang et~al.(2024{\natexlab{b}})Zhang, Liu, Xie, Zhang, Xu, and Zha}]{escnet}
Fanrui Zhang, Jiawei Liu, Jingyi Xie, Qiang Zhang, Yongchao Xu, and Zheng-Jun Zha. 2024{\natexlab{b}}.
\newblock Escnet: Entity-enhanced and stance checking network for multi-modal fact-checking.
\newblock In \emph{Proceedings of the ACM on Web Conference 2024}, pages 2429--2440.

\bibitem[{Zhang et~al.(2023{\natexlab{b}})Zhang, Liu, Zhang, Sun, Xie, and Zha}]{ecenet}
Fanrui Zhang, Jiawei Liu, Qiang Zhang, Esther Sun, Jingyi Xie, and Zheng-Jun Zha. 2023{\natexlab{b}}.
\newblock Ecenet: explainable and context-enhanced network for muti-modal fact verification.
\newblock In \emph{Proceedings of the 31st ACM International Conference on Multimedia}, pages 1231--1240.

\bibitem[{Zhang et~al.(2021)Zhang, Li, Fukumoto, and Ye}]{arsjoint}
Zhiwei Zhang, Jiyi Li, Fumiyo Fukumoto, and Yanming Ye. 2021.
\newblock Abstract, rationale, stance: A joint model for scientific claim verification.
\newblock In \emph{Proceedings of the 2021 Conference on Empirical Methods in Natural Language Processing}, pages 3580--3586.

\bibitem[{Zhao et~al.(2020)Zhao, Xiong, Rosset, Song, Bennett, and Tiwary}]{transformer_xh}
Chen Zhao, Chenyan Xiong, Corby Rosset, Xia Song, Paul Bennett, and Saurabh Tiwary. 2020.
\newblock Transformer-xh: Multi-evidence reasoning with extra hop attention.
\newblock In \emph{International Conference on Learning Representations}.

\bibitem[{Zhao et~al.(2024)Zhao, Long, Jiang, Wang, Chen, Liu, Tang, Zhang, Zhao, and Cohan}]{explainable3}
Yilun Zhao, Yitao Long, Tintin Jiang, Chengye Wang, Weiyuan Chen, Hongjun Liu, Xiangru Tang, Yiming Zhang, Chen Zhao, and Arman Cohan. 2024.
\newblock Findver: Explainable claim verification over long and hybrid-content financial documents.
\newblock In \emph{Proceedings of the 2024 Conference on Empirical Methods in Natural Language Processing}, pages 14739--14752.

\bibitem[{Zheng et~al.(2024)Zheng, Li, Zhang, Shang, Huang, and Jia}]{rav}
Liwen Zheng, Chaozhuo Li, Xi~Zhang, Yu-Ming Shang, Feiran Huang, and Haoran Jia. 2024.
\newblock Evidence retrieval is almost all you need for fact verification.
\newblock In \emph{Findings of the Association for Computational Linguistics ACL 2024}, pages 9274--9281.

\bibitem[{Zhong et~al.(2020)Zhong, Xu, Tang, Xu, Duan, Zhou, Wang, and Yin}]{dream}
Wanjun Zhong, Jingjing Xu, Duyu Tang, Zenan Xu, Nan Duan, Ming Zhou, Jiahai Wang, and Jian Yin. 2020.
\newblock Reasoning over semantic-level graph for fact checking.
\newblock In \emph{Proceedings of the 58th Annual Meeting of the Association for Computational Linguistics}, pages 6170--6180.

\bibitem[{Zhou et~al.(2019)Zhou, Han, Yang, Liu, Wang, Li, and Sun}]{gear}
Jie Zhou, Xu~Han, Cheng Yang, Zhiyuan Liu, Lifeng Wang, Changcheng Li, and Maosong Sun. 2019.
\newblock Gear: Graph-based evidence aggregating and reasoning for fact verification.
\newblock In \emph{Proceedings of the 57th Annual Meeting of the Association for Computational Linguistics}, pages 892--901.

\bibitem[{Zhou et~al.(2023)Zhou, Fung, Chen, Thomas, Ji, and Chang}]{chartt5}
Mingyang Zhou, Yi~Fung, Long Chen, Christopher Thomas, Heng Ji, and Shih-Fu Chang. 2023.
\newblock Enhanced chart understanding via visual language pre-training on plot table pairs.
\newblock In \emph{Findings of the Association for Computational Linguistics: ACL 2023}, pages 1314--1326.

\bibitem[{Zou et~al.(2023)Zou, Zhang, and Zhao}]{decker}
Anni Zou, Zhuosheng Zhang, and Hai Zhao. 2023.
\newblock Decker: Double check with heterogeneous knowledge for commonsense fact verification.
\newblock In \emph{Findings of the Association for Computational Linguistics: ACL 2023}, pages 11891--11904.

\end{thebibliography}

\clearpage
\appendix
\section{Mathematical Notations}
\label{sec:notations}

Here we provide a summary of main mathematical notations used in the main paper in Table \ref{table:notation_table}.

\begin{table*}[t]
	\centering
	\caption{Summary of mathematical notations.}
	\resizebox{\textwidth}{!}{
		\begin{tabular}{c|l}
			\toprule
			Notation  & Description  \\
			\hline
			$ \mathcal{D} $ & a claim verification dataset \\
			$ \mathcal{C} $ & a set of $ N=|\mathcal{C}| $ claims \\
			$ \mathcal{T} $ & a corpus of $ T=|\mathcal{T}| $ evidence texts \\
			$ \mathcal{I} $ & a set of $ I=|\mathcal{I}| $ images \\
                $ \mathcal{E} $ & a set of $ N=|\mathcal{C}| $ explanations for $ N $ corresponding claims \\
                $ \mathcal{Y} $ & a set of labels \\
                $ \textbf{h}_{t,\text{CLS}}^{(l)} $ & evidence text $ t $'s [CLS] token embedding at the $ l $-th step \\
                $ \textbf{z}_{i,\text{CLS}}^{(l)} $ & evidence image $ i $'s [CLS] token embedding at the $ l $-th step \\
                \hline
                $ \textbf{u}_{c,\text{CLS}} $ & unified multi-modal claim emebdding after token-level fusion, $ \textbf{c}=\textbf{u}_{c,\text{CLS}} $ \\
                $ \textbf{t} $ & unified multi-modal evidence embedding after token-level fusion \\
                $ \hat{\textbf{y}} $ & predicted label probability distribution from the verification module \\
                \hline
                $ \bar{\textbf{E}}_c $ & averaged multi-modal embedding matrix used to input to language model decoder for explanation generation \\
                $ \bm{\mathscr{L}}(e_j|\cdot) $ & $ V $-dimensional logits from language model decoder when generating token $ e_j $ \\
                $ \hat{\textbf{y}}_e $ & predicted label probability distribution from the explanation module \\
			\bottomrule
		\end{tabular}
	}
	\label{table:notation_table}
\end{table*}
\section{Pseudo-Code of Training Process}
\label{sec:algorithm}

We summarize the training process at Algorithms \ref{algo:training_algorithm1}--\ref{algo:training_algorithm2}. We first optimize multi-modal evidence retriever by Algorithm \ref{algo:training_algorithm1}. After convergence, we fix the parameters of the retriever and use it to produce retrieved evidence for claims. For claim verification and explanation generation in Algorithm \ref{algo:training_algorithm2}, we jointly optimize both objectives using retrieved evidence in Algorithm \ref{algo:training_algorithm1}.

\begin{algorithm}
	\caption{Multi-Modal Evidence Retrieval}
	\label{algo:training_algorithm1}
	\begin{flushleft}
		\hspace*{\algorithmicindent}\textbf{Input}: A claim verification dataset $ \mathcal{D} $ with claims $ \mathcal{C} $, evidence texts $ \mathcal{T} $, and images $ \mathcal{I} $. \\
		\hspace*{\algorithmicindent}\textbf{Output}: Retrieved evidence for test claims.
	\end{flushleft} 
	\begin{algorithmic}[1] 
		\State Initialize model with pre-trained parameters in scientific domain or general domain.
		\While{not converged}
        \State Construct two-layer multi-modal graph for each claim with text and images and each evidence with text and images.
            \For{evidence text $ t $ and images $ i\in\mathcal{I}(t) $}\label{algo:for_loop}
            \State Feature initialization by $ \textbf{H}_t^{(l=1)}=\text{TRM}(\textbf{H}_t^{(l=0)}) $ and $ \textbf{Z}_i^{(l=1)}=\text{ViT}(\textbf{Z}_i^{(l=0)}) $ and obtain token embeddings $ \textbf{H}_e^{(l=1)} $ for evidence sentence $ e $.
                \For{$ l=1,2,...,L-1 $}
                \Statex \qquad\qquad\textit{ // Image-to-text reasoning}
                \State Image-to-text reasoning by Eq. \ref{eq:graph_conv_layer}.
                \State Transformer step with multi-modal multi-head attention Eq. \ref{eq:multi_head_attention}
                \Statex \qquad\qquad\textit{ // Text-to-image reasoning}
                \State Text-to-image reasoning by Eq. \ref{eq:text_to_image_reasoning}.
                \State ViT step with multi-modal multi-head attention Eq. \ref{eq:multi_head_attention}.
                \EndFor
            \EndFor\label{algo:end_for}
        \State Evidence embedding $ \textbf{h}_t=\textbf{h}_{t,\text{CLS}}^{(L)} $.
        \State Repeat Lines \ref{algo:for_loop}--\ref{algo:end_for} and obtain claim embedding $ \textbf{h}_c=\textbf{h}_{c,\text{CLS}}^{(L)} $.
        \State Minimize loss function $ \mathcal{L}_{\text{Ret}} $ in Eq. \ref{eq:retrieval_objective_function} with Adam optimizer.
    \EndWhile
	\end{algorithmic}
\end{algorithm}

\begin{algorithm}
	\caption{Multi-Modal Claim Verification and Explanation Generation}
	\label{algo:training_algorithm2}
	\begin{flushleft}
		\hspace*{\algorithmicindent}\textbf{Input}: A claim verification dataset $ \mathcal{D} $ with claims $ \mathcal{C} $, evidence texts $ \mathcal{T} $, and images $ \mathcal{I} $. \\
		\hspace*{\algorithmicindent}\textbf{Output}: Predicted veracity labels and generated explanations for test claims.
	\end{flushleft} 
	\begin{algorithmic}[1] 
		\State Initialize model with pre-trained parameters in scientific domain or general domain.
            \State Use trained retriever in Algorithm \ref{algo:training_algorithm1} to output retrieved evidence for claims.
		\While{not converged}
            \State Obtain claim and evidence embedding matrices by multi-modal graph encoder Eq. \ref{eq:multi_modal_graph_encoder}.
            \For{each claim $ c $}
            \Statex \qquad\quad\textit{ // Token-level fusion}
            \State Obtain unified claim and evidence embedding matrices $ \textbf{U}_c $ and $ \textbf{U}_{t_k} $ by Eqs. \ref{eq:multi_head_attention_token_level}--\ref{eq:linear_projection_token_level}.
            \State Claim-evidence interaction and obtain $ \textbf{U}_{c,t_k} $ by Eq. \ref{eq:claim_evidence_interaction}.
            \State $ \textbf{U}_c=\text{mean}\big(\{\textbf{U}_{t_k,c}|t_k\in\mathcal{T}(c)\}\big) $ and obtain claim embedding $ \textbf{c}=\textbf{u}_{c,\text{CLS}} $.
            \Statex \qquad\quad\textit{ // Evidence-level fusion}
            \State Obtain evidence embedding $ \textbf{t} $ by Eqs. \ref{eq:image_fusion}--\ref{eq:evidence_fusion}.
            \State $ \hat{\textbf{y}}= \text{softmax}(f_{\text{MLP}}([\textbf{c}||\textbf{t}])) $.
            \Statex \qquad\quad\textit{ // Explanation generation}
            \State Obtain concatenated 
            $ \widehat{\textbf{E}}_{c,t_k} $ by Eq. \ref{eq:concatenated_embedding_matrix}.
            \State $ \bar{\textbf{E}}_{c}=\text{mean}(\tilde{\textbf{E}}_{c,t_1},...,\tilde{\textbf{E}}_{c,t_K}) $.
            \State $ \hat{e}=f_{\text{LMDec}}(\bar{\textbf{E}}_{c,t}) $.
            \Statex \qquad\quad\textit{ // Consistency regularization}
            \State For each word $ e_j $, obtain logit $ \bm{\mathscr{L}}(e_j) $.
            \State Mean pooling by Eq. \ref{eq:logits_mean_pooling}.
            \State $ \hat{\textbf{y}}_e=\text{softmax}(f_{\text{MLP}}(\bm{\mathscr{L}}) $).
            \EndFor
            \State Minimize overall loss function Eq. \ref{eq:ver_and_exp_loss}.
    \EndWhile
	\end{algorithmic}
\end{algorithm}
\section{Complexity Analysis}
\label{sec:complexity_analysis}

Here we analyze parameter count and complexity, and make a short comment on running time.

\textbf{Computational Complexity.} For evidence retrieval, we have $ \mathcal{O}(L((W+2)d^2+(W+2)^2d+d^2)) $ where $ W $ is the context length for positional encodings. For claim verification, we have $ \mathcal{O}(W^2|\mathcal{I}(t)|Kd) $ where $ |\mathcal{I}(t)| $ is the number of images each evidence text or claim text has, and $ K $ is the number of retrieved evidence. For explanation generation, we have $\mathcal{O}(W^2dLK)$. Overall, the complexity is quadratic, which is a standard complexity for many language models \cite{transformer,bert,transformer_xh,t5,clip}.

\textbf{Parameter Count.} For evidence retrieval, we have $ (12Ld^2+Vd+Wd)+(12Ld^2+P^2C+Wd)+(d^2+2d) $ parameters. $ L $ is number of language modeling steps, $ d $ is the dimension of hidden embeddings, $ V $ is the size of language model vocabulary, $ W $ is the context length for positional encodings, $ P $ is the size of each patch in ViT, and $ C $ is the number of input channels. For claim verification, we have $ 8d^2 $ parameters. For explanation generation, we have $ 28Ld^2+Vd+Wd $ parameters.

\textbf{Short Comment on Running Time.} The main focus of our paper is on model effectiveness, not running efficiency. But for completeness, we still briefly report running time. On the largest dataset MR2, our model takes around 1.5 hours to converge. Experiments were conducted on 4 NVIDIA A100 GPUs. One possible method to improve running efficiency is to replace language model with LongFormer \cite{longformer}, which is efficient and supports long texts. Another method is to use distributed and parallel training on more and larger GPUs. We leave the research of optimizing running efficiency as a future work.
\section{Additional Discussion on Model Architecture}

In this section we provide more explanations of the motivation behind model architecture to make the paper clearer and more self-contained.

\textbf{Motivation of the GNN Module.} For GNN, it supports variable number of neighbors for each node, and the main function of GNN is to aggregate neighboring nodes into a unified embedding. In our scenario, a text is associated with variable number of images. Although there are 5 or fewer images for each text, GNN can effectively aggregate images into a unified image embedding for language and vision modeling. Our design is similar to GEAR \cite{gear}, which also uses GNN to aggregate 5 or fewer evidence texts for fact-checking. Both GEAR and our experiments verify that GNN is an effective module for small-sized evidence graph. Our evidence graph is a multi-modal heterogeneous graph with both texts and images, and we use different matrices to project text embeddings and image embeddings to the same embedding space for GNN aggregation, please see Eq. \ref{eq:linear_projection} where we use $ \textbf{W}_{\text{txt}} $ and $ \textbf{W}_{\text{img}} $ for text and image, respectively.

\textbf{Explanation on the Size of Multi-Modal Evidence Graph.} Both our work and GEAR process an evidence graph with 5 or fewer nodes, and both papers show the effectiveness of applying GNN to process such an evidence graph. The additional ablation study provided above also show that GNN aggregator is more effective than mean and max pooling aggregator. We would like to clarify that the purpose of GNN does not lie in how many neighbors a node has, but instead, the purpose lies in neighboring node aggregation with variable number of neighbors. The widely used Cora citation graph \cite{gat} in graph community has 4 neighbors for each node on average, and GNNs are effective at learning node embedding on such a graph. Furthermore, as shown in GraphSAGE \cite{graphsage}, the neighbor sampling strategy can sample as few as 2 neighbors for effective neighbor aggregation, and our work also uses the same neighbor sampling strategy to aggregate images of a text.

\textbf{Motivation of Toekn-level and Evidence-level Fusion.} For token-level and evidence-level fusion, the multi-modal graph encoder outputs embedding matrices for claims and evidence separately, and there is no interaction or reasoning between claim and evidence. We aim to use token- and evidence-level fusion to allow multi-modal claim and evidence to interact and reason with each other for claim verification.

\textbf{Motivation of Multi-Modal Fusion-in-Decoder.} For multi-modal fusion-in-decoder, we aim to aggregate multiple evidence texts with their images for explanation generation. Multi-modal fusion-in-decoder can effectively achieve this goal by concatenating image embedding with its corresponding evidence text embedding matrix, and by taking mean pooling for multiple pieces of evidence. The motivation of concatenation with image embedding is that we can incorporate image information into explanation generation. The motivation of taking mean pooling is that we effectively use the information of multiple evidence for explanation generation. Concatenating multiple evidence but not taking mean pooling is also possible, but it results in extremely long sequence, which is inefficient. Thus we choose mean pooling.

\textbf{Motivation of Consistency Regularizer.} For consistency regularizer, the motivation is similar to the design in MochegModel. We aim to make the generated explanation consistently reflect the predicted veracity, so that the explanation can well justify the reasoning process of how the model predicts the veracity.
\section{Additional Dataset Details}
\label{sec:additional_dataset_details}

Here we provide dataset creation details of our introduced AIChartClaim, as well as data preprocessing details of other three publicly available datasets.

\subsection{Details of AIChartClaim}

\textbf{Data source.} For each paper, our annotators record publication year, venue, title, and caption of the chart. If the chart is not clearly readable, annotators search for other papers with appropriate charts. Some claims in the paper usually have a prefix, e.g., ``Figure 1 shows that ...''. To make the claims self-contained, our annotators remove this prefix and retain the main scientific findings in the sentence. Since the whole dataset is created by three different annotators, we have the fourth annotator to double check the whole dataset to make the creation consistent across all the claims and evidence.

\textbf{Data augmentation.} To create negations for the claims in the paper, we follow \citet{scifact} and ask annotators to write negation for the 300 claims, taking precautions not to bias the negation by using obvious keywords, like ``not''. We mainly focus on the negation of semantic meaning of the claims. The ``claim-only'' version of our model, which removes both evidence texts and images for verification, achieves around 53\%, which is quite close to 50\%, suggesting that the negation process does not introduce severe artifacts.

To further augment the dataset by prompting GPT-4o for generation, for each chart with caption, we use below prompt to generate two more claims, one supported and one refuted by the chart:

\begin{tcolorbox}[fontupper=\small, colback=gray!5, colframe=gray!80!black, title=Prompt for Claim Augmentation]

Please provide two more claims based on the chart and its caption that require multi-step reasoning, with one supported by the information in the chart and the other refuted by the information in the chart. The new claims should be different from previous claims. Meanwhile, please provide concise explanations in less than 100 words.

\end{tcolorbox}

After generation, our annotators manually check and analyze the generated claims and explanations to make sure they are indeed correct, high-quality, and consistent with the charts. If there is any unclear or erroneous description, we either prompt GPT-4o to generate more claims with explanations, or manually correct the generated texts.

\textbf{Explanation.} We further generate explanations for those natural claims (i.e., claims obtained from papers and their negations). For each chart with caption, we use below prompt for generation:

\begin{tcolorbox}[fontupper=\small, colback=gray!5, colframe=gray!80!black, title=Prompt for Explanation Generation]

The caption of this chart is ``[CAPTION]''. There is a claim based on the chart ``[CLAIM TEXT]''. Concisely explain why this claim is [SUPPORTED / REFUTED] by the chart in less than 100 words.

\end{tcolorbox}

Here we replace [CAPTION] with the caption of the input chart image, and [CLAIM TEXT] with the claim text. We select either ``SUPPORT'' or ``REFUTE'' based on the label of the claim. Again, our annotators double check and analyze the generated explanation, and correct or regenerate it if there is any unclear or wrong description.

We split the created dataset into training:development:test by 70:10:20. For each publication venue, we randomly select 70\% claims for training, 10\% for validation, and 20\% for testing. The split is balanced, and there is no overlap between training charts and test charts.

\textbf{Human Evaluation.} To further evaluate the quality of out dataset, we follow SciFact \cite{scifact} and randomly select 100 claim-chart pairs for re-annotation for three independent times. For each time, we randomly select different 100 claim-chart pairs and ask a different PhD student specializing in AI for re-annotation. These three PhD students were never involved in dataset creation before. The label agreement is 0.72 Cohen’s $ \kappa $ \cite{scifact}, which is a high agreement and is similar to the result in SciFact.

\textbf{Motivation of Proposing AIChartClaim Dataset.} We are motivated to create AIChartClaim dataset for three main reasons. \emph{First}, most existing multi-modal fact-checking datasets are in the general domain. To our knowledge, our AIChartClaim dataset is the first scientific multi-modal fact-checking dataset in AI domain. Thus our dataset complements the research community. \emph{Second}, understanding increases and decreases in quantities in scientific charts with AI domain-specific language is a crucial scientific reasoning ability for language models, thus it is necessary to have this scientific dataset. \emph{Third}, we also make a comparison to other related datasets in Table \ref{table:dataset_comparison}. Existing scientific fact-checking datasets (SciFact, BearFact, and Check-COVID) are based on texts only, with no images. Existing multi-modal fact-checking chart datasets (ChartCheck and ChartFC) are in the general domain, and their content does not discuss scientific or more specifically, AI concepts. Other multi-modal fact-checking datasets (Mocheg, MR2, NewsCLIPpings, and FACTIFY) are general-domain datasets. They are neither chart nor scientific datasets. Such comparison shows that it is necessary to have our dataset to complement the research community.

\subsection{Details of Other Datasets}

\begin{table*}[t]
	\centering
	\caption{Result of evidence retrieval with \emph{Precision} score (\%).}
	\resizebox{\textwidth}{!}{
		\begin{tabular}{c|ccc|ccc|ccc|ccc}
			\toprule
                \multirow{2}{*}{Model} & \multicolumn{3}{c|}{AIChartClaim} & \multicolumn{3}{c|}{ChartCheck} & \multicolumn{3}{c|}{Mocheg} & \multicolumn{3}{c}{MR2} \\
			\cline{2-13}
			{} & Prec@1 & Prec@5 & Prec@7 & Prec@1 & Prec@5 & Prec@7 & Prec@1 & Prec@5 & Prec@7 & Prec@1 & Prec@5 & Prec@7 \\
			\hline
                BM25 & 43.3$ \pm $0.0 & 10.8$ \pm $0.0 & 8.3$ \pm $0.0 & 35.9$ \pm $0.0 & 9.4$ \pm $0.0 & 6.9$ \pm $0.0 & 39.6$ \pm $0.0 & 14.8$ \pm $0.0 & 11.4$ \pm $0.0 & 22.9$ \pm $0.0 & 13.5$ \pm $0.0 & 10.9$ \pm $0.0 \\
                RAV & 50.6$ \pm $1.1 & 13.5$ \pm $0.1 & 9.9$ \pm $0.1 & 53.6$ \pm $0.2 & 13.8$ \pm $0.1 & 10.3$ \pm $0.1 & 49.7$ \pm $0.6 & 20.8$ \pm $0.3 & 16.1$ \pm $0.3 & 29.6$ \pm $0.3 & 17.4$ \pm $0.3 & 14.2$ \pm $0.2 \\
                JustiLM & 53.6$ \pm $2.7 & 14.3$ \pm $0.2 & 10.7$ \pm $0.1 & 51.4$ \pm $0.7 & 13.6$ \pm $0.0 & 10.1$ \pm $0.0 & 50.3$ \pm $0.7 & 21.1$ \pm $0.2 & 16.4$ \pm $0.2 & 27.0$ \pm $0.7 & 16.4$ \pm $0.2 & 13.5$ \pm $0.1 \\
                \hline
                MochegModel & 58.8$ \pm $0.0 & 14.9$ \pm $0.0 & 11.0$ \pm $0.0 & 51.6$ \pm $0.1 & 13.3$ \pm $0.0 & 9.9$ \pm $0.0 & 41.8$ \pm $0.0 & 11.6$ \pm $0.0 & 8.8$ \pm $0.0 & 29.5$ \pm $0.1 & 19.1$ \pm $0.0 & 15.8$ \pm $0.0 \\
                TransXH+ViT & 53.3$ \pm $3.5 & 14.2$ \pm $0.5 & 10.6$ \pm $0.2 & 53.2$ \pm $0.2 & 13.8$ \pm $0.2 & 10.3$ \pm $0.1 & 51.1$ \pm $0.7 & 21.1$ \pm $0.2 & 16.4$ \pm $0.2 & 35.5$ \pm $1.2 & 22.9$ \pm $1.0 & 18.3$ \pm $0.9 \\
                \hline
                MEVER w/o images & 63.4$ \pm $1.1 & 15.3$ \pm $0.2 & 11.2$ \pm $0.0 & 51.8$ \pm $0.8 & 13.6$ \pm $0.2 & 10.1$ \pm $0.1 & 49.2$ \pm $3.9 & 20.4$ \pm $1.5 & 15.9$ \pm $1.1 & 28.2$ \pm $0.2 & 17.2$ \pm $0.2 & 14.2$ \pm $0.0 \\
                \hline
                MEVER (ours) & \textbf{65.7$ \pm $0.6} & \textbf{15.4$ \pm $0.0} & \textbf{12.0$ \pm $1.2} & \textbf{56.0$ \pm $0.4} & \textbf{14.4$ \pm $0.1} & \textbf{10.7$ \pm $0.1} & \textbf{53.1$ \pm $0.8} & \textbf{22.0$ \pm $0.2} & \textbf{17.0$ \pm $0.2} & \textbf{37.6$ \pm $0.6} & \textbf{24.1$ \pm $0.6} & \textbf{19.3$ \pm $0.1} \\
			\bottomrule
		\end{tabular}
	}
	\label{table:precision}
\end{table*}

\begin{table*}[t]
	\centering
	\caption{Result of evidence retrieval with \emph{Recall} score (\%).}
	\resizebox{\textwidth}{!}{
		\begin{tabular}{c|ccc|ccc|ccc|ccc}
			\toprule
                \multirow{2}{*}{Model} & \multicolumn{3}{c|}{AIChartClaim} & \multicolumn{3}{c|}{ChartCheck} & \multicolumn{3}{c|}{Mocheg} & \multicolumn{3}{c}{MR2} \\
			\cline{2-13}
			{} & Rec@1 & Rec@5 & Rec@7 & Rec@1 & Rec@5 & Rec@7 & Rec@1 & Rec@5 & Rec@7 & Rec@1 & Rec@5 & Rec@7 \\
			\hline
                BM25 & 43.3$ \pm $0.0 & 54.2$ \pm $0.0 & 57.9$ \pm $0.0 & 35.9$ \pm $0.0 & 46.9$ \pm $0.0 & 48.5$ \pm $0.0 & 17.9$ \pm $0.0 & 30.8$ \pm $0.0 & 32.9$ \pm $0.0 & 3.8$ \pm $0.0 & 10.6$ \pm $0.0 & 11.9$ \pm $0.0 \\
                RAV & 50.6$ \pm $1.1 & 67.4$ \pm $0.6 & 69.4$ \pm $1.0 & 53.6$ \pm $0.2 & 69.2$ \pm $0.5 & 72.0$ \pm $0.3 & 22.8$ \pm $0.4 & 43.0$ \pm $0.7 & 46.4$ \pm $0.8 & 5.0$ \pm $0.1 & 13.8$ \pm $0.3 & 15.7$ \pm $0.4 \\
                JustiLM & 53.6$ \pm $2.7 & 71.5$ \pm $0.9 & 74.6$ \pm $0.4 & 51.4$ \pm $0.7 & 67.8$ \pm $0.1 & 70.7$ \pm $0.2 & 23.4$ \pm $0.3 & 43.8$ \pm $0.5 & 47.0$ \pm $0.5 & 4.5$ \pm $0.1 & 13.1$ \pm $0.1 & 14.9$ \pm $0.1 \\
                \hline
                MochegModel & 58.8$ \pm $0.0 & 74.6$ \pm $0.0 & 77.1$ \pm $0.0 & 51.6$ \pm $0.1 & 66.7$ \pm $0.0 & 69.1$ \pm $0.1 & 22.2$ \pm $0.0 & 30.2$ \pm $0.0 & 31.6$ \pm $0.0 & 4.9$ \pm $0.0 & 14.9$ \pm $0.0 & 16.9$ \pm $0.0 \\
                TransXH+ViT & 53.3$ \pm $3.5 & 71.0$ \pm $2.7 & 74.0$ \pm $1.5 & 53.2$ \pm $0.2 & 69.0$ \pm $0.8 & 71.8$ \pm $0.4 & 23.4$ \pm $0.4 & 30.2$ \pm $0.0 & 31.6$ \pm $0.0 & 5.6$ \pm $0.2 & 16.5$ \pm $0.6 & 18.4$ \pm $0.8 \\
                \hline
                MEVER w/o images & 63.4$ \pm $1.1 & 76.5$ \pm $0.9 & 78.2$ \pm $0.2 & 51.8$ \pm $0.8 & 67.8$ \pm $0.9 & 70.5$ \pm $0.9 & 22.6$ \pm $1.7 & 42.3$ \pm $2.1 & 45.6$ \pm $3.2 & 4.8$ \pm $0.0 & 13.7$ \pm $0.0 & 15.7$ \pm $0.1 \\
                \hline
                MEVER (ours) & \textbf{65.7$ \pm $0.6} & \textbf{77.2$ \pm $0.2} & \textbf{79.0$ \pm $0.5} & \textbf{56.0$ \pm $0.4} & \textbf{72.2$ \pm $0.5} & \textbf{75.0$ \pm $0.5} & \textbf{24.4$ \pm $0.4} & \textbf{45.2$ \pm $0.4} & \textbf{48.7$ \pm $0.4} & \textbf{5.9$ \pm $0.1} & \textbf{17.3$ \pm $0.3} & \textbf{19.2$ \pm $0.4} \\
			\bottomrule
		\end{tabular}
	}
	\label{table:recall}
\end{table*}

\textbf{ChartCheck}\footnote{\url{https://github.com/mubasharaak/ChartCheck}} is a chart dataset in general domain with chart images and textual captions as multi-modal evidence. It also has explanations. Since the chart images in the dataset are provided by URL links, we accordingly download the charts using the links and remove charts and associated claims if the links are unavailable. In total, we have 1,615 charts with captions and 10,038 claims.

\textbf{Mocheg}\footnote{\url{https://github.com/VT-NLP/Mocheg}} is a multi-modal dataset in general domain with explanations. We follow the instructions in GitHub and download the dataset. However, in their original paper \cite{mocheg}, authors respectively use three different preprocessing methods for evidence retrieval, claim verification, and explanation generation, resulting in three different variations or subsets of the original dataset. In our model, we aim to keep consistent across different tasks, thus we keep all the claims, evidence, and images, and consistently use the same set of data for all three tasks. This is why our results, especially explanation generation, are slightly different from the results in the original paper, which uses only half of the claims for explanation generation. Finally, since some evidence texts do not have associated images, we use pre-trianed CLIP \cite{clip} to align texts and images. For each evidence text, we select top-3 most similar images and associate them with the text.

\textbf{MR2}\footnote{\url{https://github.com/THU-BPM/MR2}} is another multi-modal dataset in general domain, but with no explanations. Both its claims and evidence texts have associated images. The downloaded dataset using the link in GitHub is slightly different from the one reported in the original paper in terms of dataset size \cite{mr2}. We tried very hard but still cannot obtain the same dataset. Thus, the results in our paper also have some deviations from the results in their original paper. The images of some evidence texts are provided in the dataset, while the images of other evidence texts are provided by a URL link. We accordingly download the images based on the link, and remove evidence text whose image link is unavailable. In total, we have 13,785 claims, 91,347 evidence texts, and 105,132 images.



\section{Experiment Environment}
\label{sec:experiment_environment}

All the experiments were conducted on Linux server with 4 NVIDIA A100-SXM4-80GB GPUs. Its operating system is 20.04.5 LTS (Focal Fossa). We implemented our proposed model MEVER using Python 3.9 as programming language and PyTorch 2.4.1 as deep learning library. Other frameworks include numpy 1.24.1, sklearn 1.3.2, and transformers 4.46.0.
\section{Additional Experiment Results}
\label{sec:additional_experiments}

Here we provide additional experiment results in terms of multi-modal evidence retrieval, claim verification, and explanation generation.

\subsection{Evidence Retrieval}
\label{sec:additional_evidence_retrieval}

In the main paper, we report results of MAP, Precision@$ \kappa $, and Recall@$ \kappa $ ($ \kappa=3 $) for evidence retrieval. Here we provide additional results of precision and recall scores when varying $ \kappa $ in \{1, 5, 7\} in Tables \ref{table:precision}--\ref{table:recall}. Similarly, multi-modal baselines tend to outperform text-only baselines, verifying that images indeed bring useful information to improve the retrieval performance. Our model further outperforms multi-modal baselines, since we design a nested architecture to well integrate graph reasoning and vision-language modeling, thereby improving the result. The ablated version of our model, which removes images, deteriorates the retrieval performance, which further demonstrates that our model effectively incorporates images for multi-modal retrieval.

\subsection{Claim Verification}
\label{sec:additional_claim_verification}

In the main paper we report claim verification results with Macro F1 score. Here we provide additional results with Micro F1 score in Table \ref{table:claim_verification_micro_f1}. Overall, multi-modal baseline models tend to outperform text-only baselines, since images provide auxiliary information to boost the verification accuracy. Our model further produces higher accuracy, since we model both token-level and evidence-level fusion to well reason between multi-modal claims and evidence, thereby achieving a more accurate verification. Though GPT-4o is slightly better than our model on the general-domain Mocheg dataset, our model still produces better results on Macro F1 score and on other datasets.


\subsection{Explanation Generation}
\label{sec:additional_explanation_generation}

We present ROUGE-L, METEOR, and BLEU-2 scores with retrieved setting in the paper for explanation generation. Here we further provide ROUGE-1, ROUGE-2, and BLEU-4 scores with retrieved setting in Table \ref{table:explanation_generation2}. Similarly, we show the results of explanation generation with gold setting in Tables \ref{table:explanation_generation3}--\ref{table:explanation_generation4}. Specifically, Table \ref{table:explanation_generation2} presents additional evaluation metrics, i.e., ROUGE-1, ROUGE-2, and BLEU-4, for explanation generation with retrieved setting. Tables \ref{table:explanation_generation3}--\ref{table:explanation_generation4} show all six metrics of explanation generation with gold setting. Overall, we observe that multi-modal baselines generate explanations more accurately than text-only baselines, since images complement texts for high-quality generation. Our model also outperforms baselines in most cases, showcasing the advantage of our multi-modal Fusion-in-Decoder and consistency regularizer. 

To comprehensively evaluate explanation generation, we further adopt VLM-as-a-Judge for evaluation. Following existing works \cite{vlm_as_a_judge}, we input below prompt to LLaVA-v1.5-7B \cite{llava,llava2}:

\begin{tcolorbox}[fontupper=\small, colback=gray!5, colframe=gray!80!black, title=Prompt for VLM-as-a-Judge]

USER: <image> You are an expert evaluator for fact-checking explanations. Please evaluate the quality of the following explanation, based on the provided claim, evidence text, evidence image, and predicted label. Your evaluation should consider the following four factors:\\

1. Label-Explanation Consistency: Does the explanation appropriately justify the predicted label?

2. Relevance: Is the explanation relevant to the claim, evidence text, and evidence image?

3. Correctness: Is the explanation factually correct based on the evidence text and evidence image?

4. Clarity: Is the explanation clear and easy to understand?\\

For each factor, assign a score from 1 (poor) to 5 (excellent).\\ 

Claim: [CLAIM TEXT]

Evidence Text: [EVIDENCE TEXT]

Predicted Label: [PREDICTED LABEL]

Explanation: [GENERATED EXPLANATION]\\

Please output your answer in the following format:

Label-Explanation Consistency: [SCORE]

Relevance: [SCORE]

Correctness: [SCORE]

Clarity: [SCORE]

ASSISTANT:
\end{tcolorbox}

Table \ref{table:vlm_as_a_judge} shows the results. Overall, our model outperforms baselines in most cases. Although GPT-4o performs slightly better than our model for Correctness and Clarity on Mocheg dataset, our model still significantly produces better explanation on other evaluation dimensions and datasets. UniChart and ChartGemma are specifically designed for chart data, thus we do not show their result on Mocheg dataset.

\subsection{Further Discussion on Three Evaluation Tasks}

We use three tasks with multiple metrics to evaluate the performance of our model. There is no baseline model that performs very well on all tasks and metrics, while our model consistently performs better than or at least comparably with the best baseline on all tasks and metrics.

Specifically, for claim verification in Table \ref{table:claim_verification_macro_f1}, ChartGemma and Transformer-XH++ perform well on ChartCheck dataset, but on Mocheg and MR2 datasets Transformer-XH++ significantly deteriorates the performance (we conduct paired t-test with $ p=0.05 $), and ChartGemma is even not designed for non-chart datasets. Similarly, GPT-4o and KGAT produce good result on Mocheg dataset, but on other datasets they fall behind our model with statistical significance (we conduct paired t-test with $ p=0.05 $). This indicates that our model at least does not hurt the performance of any individual task or metric, but can significantly outperform baselines on many other tasks and metrics.

Similarly, for explanation generation, ChartGemma and ECENet produce good explanation on ChartCheck, but the performance of ECENet on Mocheg dataset is statistically significantly lower than our model’s (again, ChartGemma is even not proposed for non-chart dataset, Mocheg). DePlot+FlanT5 performs well on Mocheg dataset, but on other datasets our model outperforms it with statistical significance. These results again verify that our model not only achieves better or comparable results on baselines’ respective advantageous datasets, but significantly improves them on many other datasets, tasks, and metrics.

\begin{table*}[t]
	\centering
	\caption{Explanation generation with \emph{ROUGE-1}, \emph{ROUGE-2}, and \emph{BLEU-4} scores (\%) with \emph{retrieved evidence setting}.}
	\resizebox{\textwidth}{!}{
		\begin{tabular}{c|ccc|ccc|ccc}
			\toprule
                \multirow{2}{*}{Model} & \multicolumn{3}{c|}{AIChartClaim} & \multicolumn{3}{c|}{ChartCheck} & \multicolumn{3}{c}{Mocheg} \\
			\cline{2-10}
			{} & ROUGE-1 & ROUGE-2 & BLEU-4 & ROUGE-1 & ROUGE-2 & BLEU-4 & ROUGE-1 & ROUGE-2 & BLEU-4 \\
			\hline
                JustiLM & 26.8$ \pm $0.9 & 11.5$ \pm $0.6 & 5.6$ \pm $0.5 & 41.1$ \pm $2.2 & 22.7$ \pm $1.1 & 14.4$ \pm $1.1 & 25.1$ \pm $0.3 & 11.3$ \pm $0.4 & 8.2$ \pm $0.6 \\
                \hline
                MochegModel & 41.5$ \pm $1.2 & 20.5$ \pm $1.3 & 9.4$ \pm $1.1 & 47.1$ \pm $0.4 & 26.6$ \pm $0.4 & 15.0$ \pm $0.3 & 26.0$ \pm $0.1 & 11.0$ \pm $0.5 & \textbf{10.7$ \pm $0.1} \\
                ECENet & 40.9$ \pm $0.6 & 20.5$ \pm $0.8 & 10.6$ \pm $0.5 & 46.9$ \pm $0.3 & 26.0$ \pm $0.2 & 15.1$ \pm $0.2 & 24.9$ \pm $0.4 & 11.1$ \pm $0.5 & 8.1$ \pm $0.2 \\
                DePlot+FlanT5 & 41.6$ \pm $1.3 & 20.6$ \pm $1.3 & 9.5$ \pm $1.2 & 47.3$ \pm $0.2 & 26.4$ \pm $0.1 & 15.2$ \pm $0.1 & 27.4$ \pm $0.8 & 11.8$ \pm $0.3 & 9.9$ \pm $0.4 \\
                GPT-4o & 24.1$ \pm $0.6 & 18.4$ \pm $0.5 & 7.1$ \pm $0.2 & 22.5$ \pm $0.2 & 8.9$ \pm $0.6 & 12.7$ \pm $0.2 & \textbf{30.3$ \pm $1.7} & 9.9$ \pm $1.7 & 4.5$ \pm $0.8 \\
                \hline
                UniChart & 41.3$ \pm $0.7 & 20.7$ \pm $0.9 & 10.7$ \pm $0.7 & 47.7$ \pm $0.3 & 26.5$ \pm $0.3 & 15.7$ \pm $0.1 & N.A. & N.A. & N.A. \\
                ChartGemma & 41.9$ \pm $0.2 & 21.3$ \pm $0.3 & 11.0$ \pm $0.1 & 48.0$ \pm $0.0 & 26.7$ \pm $0.2 & 15.1$ \pm $0.2 & N.A. & N.A. & N.A. \\
                \hline
                MEVER w/o images & 41.9$ \pm $0.2 & 21.6$ \pm $0.4 & 11.7$ \pm $0.1 & 47.5$ \pm $0.0 & 26.5$ \pm $0.4 & 15.8$ \pm $0.2 & 28.0$ \pm $0.2 & 13.7$ \pm $0.1 & 9.1$ \pm $0.1 \\
                \hline
                MEVER (ours) & \textbf{42.7$ \pm $0.3} & \textbf{22.0$ \pm $0.3} & \textbf{11.8$ \pm $0.4} & \textbf{48.7$ \pm $0.1} & \textbf{27.2$ \pm $0.1} & \textbf{16.8$ \pm $0.2} & 28.5$ \pm $0.3 & \textbf{14.3$ \pm $0.2} & 10.1$ \pm $0.2 \\
			\bottomrule
		\end{tabular}
	}
	\label{table:explanation_generation2}
\end{table*}

\begin{table*}[t]
	\centering
	\caption{Explanation generation with \emph{ROUGE-L}, \emph{METEOR}, and \emph{BLEU-2} scores (\%) with \emph{gold evidence setting}.}
	\resizebox{\textwidth}{!}{
		\begin{tabular}{c|ccc|ccc|ccc}
			\toprule
                \multirow{2}{*}{Model} & \multicolumn{3}{c|}{AIChartClaim} & \multicolumn{3}{c|}{ChartCheck} & \multicolumn{3}{c}{Mocheg} \\
			\cline{2-10}
			{} & ROUGE-L & METEOR & BLEU-2 & ROUGE-L & METEOR & BLEU-2 & ROUGE-L & METEOR & BLEU-2 \\
			\hline
                JustiLM & 25.6$ \pm $0.6 & 19.9$ \pm $0.8 & 14.8$ \pm $0.8 & 34.3$ \pm $0.8 & 30.6$ \pm $0.9 & 23.9$ \pm $0.7 & 23.3$ \pm $0.3 & 22.2$ \pm $0.6 & 17.6$ \pm $1.4 \\
                \hline
                MochegModel & 32.0$ \pm $1.1 & 23.8$ \pm $1.3 & 17.6$ \pm $1.3 & 38.4$ \pm $0.8 & 33.1$ \pm $0.8 & 25.0$ \pm $0.9 & 23.2$ \pm $0.1 & \textbf{22.7$ \pm $0.2} & 20.1$ \pm $0.3 \\
                ECENet & 33.1$ \pm $0.2 & 26.0$ \pm $0.1 & 19.1$ \pm $0.2 & 40.0$ \pm $0.3 & 34.2$ \pm $0.2 & 25.7$ \pm $0.2 & 22.4$ \pm $0.2 & 21.3$ \pm $0.4 & 14.6$ \pm $0.2 \\
                DePlot+FlanT5 & 32.2$ \pm $0.9 & 24.0$ \pm $1.1 & 17.6$ \pm $1.3 & 38.7$ \pm $0.8 & 33.1$ \pm $0.7 & 24.7$ \pm $0.7 & 23.2$ \pm $0.1 & 23.1$ \pm $0.1 & \textbf{20.6$ \pm $0.2} \\
                GPT-4o & 18.7$ \pm $0.4 & 23.4$ \pm $0.3 & 16.1$ \pm $0.5 & 18.4$ \pm $0.7 & 28.5$ \pm $0.7 & 16.9$ \pm $2.3 & 17.8$ \pm $0.7 & 18.8$ \pm $0.8 & 11.7$ \pm $0.4 \\
                \hline
                UniChart & 32.5$ \pm $0.9 & 25.6$ \pm $0.5 & 19.1$ \pm $0.4 & 40.1$ \pm $0.3 & 34.5$ \pm $0.5 & 26.4$ \pm $0.3 & N.A. & N.A. & N.A. \\
                ChartGemma & 33.7$ \pm $0.2 & 26.7$ \pm $0.1 & 19.7$ \pm $0.3 & 40.0$ \pm $0.2 & 34.7$ \pm $0.1 & 25.7$ \pm $0.2 & N.A. & N.A. & N.A. \\
                \hline
                MEVER w/o images & 33.5$ \pm $0.1 & 26.2$ \pm $0.1 & 20.6$ \pm $0.1 & 39.8$ \pm $0.2 & 35.4$ \pm $0.4 & 26.6$ \pm $0.4 & 23.7$ \pm $1.1 & 21.8$ \pm $0.2 & 15.0$ \pm $0.2 \\
                \hline
                MEVER (ours) & \textbf{34.4$ \pm $0.1} & \textbf{27.5$ \pm $0.2} & \textbf{21.2$ \pm $0.3} & \textbf{40.8$ \pm $0.1} & \textbf{35.9$ \pm $0.2} & \textbf{27.1$ \pm $0.3} & \textbf{24.5$ \pm $0.9} & \textbf{22.2$ \pm $0.9} & 16.6$ \pm $0.2 \\
			\bottomrule
		\end{tabular}
	}
	\label{table:explanation_generation3}
\end{table*}

\begin{table*}[t]
	\centering
	\caption{Explanation generation with \emph{ROUGE-1}, \emph{ROUGE-2}, and \emph{BLEU-4} scores (\%) with \emph{gold evidence setting}.}
	\resizebox{\textwidth}{!}{
		\begin{tabular}{c|ccc|ccc|ccc}
			\toprule
                \multirow{2}{*}{Model} & \multicolumn{3}{c|}{AIChartClaim} & \multicolumn{3}{c|}{ChartCheck} & \multicolumn{3}{c}{Mocheg} \\
			\cline{2-10}
			{} & ROUGE-1 & ROUGE-2 & BLEU-4 & ROUGE-1 & ROUGE-2 & BLEU-4 & ROUGE-1 & ROUGE-2 & BLEU-4 \\
			\hline
                JustiLM & 31.4$ \pm $0.9 & 14.0$ \pm $0.6 & 7.0$ \pm $0.4 & 40.0$ \pm $2.5 & 22.5$ \pm $0.2 & 14.4$ \pm $0.5 & 30.7$ \pm $0.8 & 15.8$ \pm $0.2 & 12.9$ \pm $1.1 \\
                \hline
                MochegModel & 39.5$ \pm $1.4 & 19.3$ \pm $1.1 & 8.9$ \pm $1.0 & 45.3$ \pm $1.0 & 25.4$ \pm $0.5 & 15.6$ \pm $0.6 & 31.4$ \pm $0.5 & 15.4$ \pm $0.2 & 14.4$ \pm $0.2 \\
                ECENet & 41.5$ \pm $0.2 & 20.4$ \pm $0.1 & 10.2$ \pm $0.1 & 47.4$ \pm $0.3 & 26.4$ \pm $0.4 & 15.2$ \pm $0.1 & 29.2$ \pm $0.3 & 15.2$ \pm $0.1 & 12.4$ \pm $0.1 \\
                DePlot+FlanT5 & 39.7$ \pm $1.2 & 19.4$ \pm $1.1 & 8.9$ \pm $1.0 & 45.4$ \pm $0.8 & 25.6$ \pm $0.6 & 15.4$ \pm $0.4 & \textbf{32.1$ \pm $0.2} & 15.5$ \pm $0.1 & \textbf{14.7$ \pm $0.1} \\
                GPT-4o & 25.5$ \pm $1.0 & 18.1$ \pm $0.8 & 6.8$ \pm $0.2 & 23.0$ \pm $1.2 & 9.6$ \pm $0.0 & 13.2$ \pm $0.1 & 31.2$ \pm $2.5 & 10.1$ \pm $0.5 & 4.7$ \pm $0.1 \\
                \hline
                UniChart & 41.6$ \pm $0.3 & 20.9$ \pm $0.1 & 11.0$ \pm $0.1 & 47.6$ \pm $0.2 & 26.5$ \pm $0.1 & 15.6$ \pm $0.1 & N.A. & N.A. & N.A. \\
                ChartGemma & 42.2$ \pm $0.1 & 21.3$ \pm $0.1 & 10.4$ \pm $0.1 & 48.0$ \pm $0.1 & 26.6$ \pm $0.1 & 16.6$ \pm $0.0 & N.A. & N.A. & N.A. \\
                \hline
                MEVER w/o images & 41.8$ \pm $0.2 & 20.8$ \pm $0.2 & 10.6$ \pm $0.1 & 47.6$ \pm $0.1 & 26.5$ \pm $0.4 & 16.7$ \pm $0.0 & 29.6$ \pm $1.8 & 15.8$ \pm $0.5 & 11.1$ \pm $1.4 \\
                \hline
                MEVER (ours) & \textbf{42.9$ \pm $0.1} & \textbf{21.9$ \pm $0.1} & \textbf{11.6$ \pm $0.2} & \textbf{48.9$ \pm $0.0} & \textbf{27.4$ \pm $0.2} & \textbf{16.8$ \pm $0.1} & \textbf{31.4$ \pm $1.2} & \textbf{16.4$ \pm $0.7} & 12.4$ \pm $0.9 \\
			\bottomrule
		\end{tabular}
	}
	\label{table:explanation_generation4}
\end{table*}

\begin{table*}[t]
	\centering
	\caption{Explanation generation with \emph{VLM-as-a-Judge}.}
	\resizebox{\textwidth}{!}{
		\begin{tabular}{c|cccc|cccc|cccc}
			\toprule
                \multirow{2}{*}{Model} & \multicolumn{4}{c|}{AIChartClaim} & \multicolumn{4}{c|}{ChartCheck} & \multicolumn{4}{c}{Mocheg} \\
			\cline{2-13}
			{} & Consistency & Relevance & Correctness & Clarity & Consistency & Relevance & Correctness & Clarity & Consistency & Relevance & Correctness & Clarity \\
			\hline
                JustiLM & 4.38$ \pm $0.05 & 4.45$ \pm $0.04 & 4.18$ \pm $0.02 & 4.34$ \pm $0.03 & 4.40$ \pm $0.02 & 4.49$ \pm $0.04 & 4.29$ \pm $0.04 & 4.43$ \pm $0.02 & 4.01$ \pm $0.02 & 4.13$ \pm $0.02 & 3.61$ \pm $0.01 & 3.86$ \pm $0.02 \\
                \hline
                MochegModel & 4.41$ \pm $0.04 & 4.51$ \pm $0.03 & 4.29$ \pm $0.04 & 4.44$ \pm $0.03 & 4.45$ \pm $0.03 & 4.55$ \pm $0.04 & 4.32$ \pm $0.02 & 4.45$ \pm $0.04 & 4.09$ \pm $0.01 & 4.24$ \pm $0.01 & 3.64$ \pm $0.04 & 3.91$ \pm $0.01 \\
                ECENet & 4.42$ \pm $0.02 & 4.52$ \pm $0.02 & 4.28$ \pm $0.03 & 4.41$ \pm $0.02 & 4.44$ \pm $0.05 & 4.54$ \pm $0.04 & 4.30$ \pm $0.03 & 4.44$ \pm $0.04 & 4.09$ \pm $0.01 & 4.23$ \pm $0.01 & 3.64$ \pm $0.02 & 3.90$ \pm $0.01 \\
                DePlot+FlanT5 & 4.44$ \pm $0.04 & 4.52$ \pm $0.04 & 4.28$ \pm $0.02 & 4.42$ \pm $0.02 & 4.43$ \pm $0.01 & 4.53$ \pm $0.01 & 4.29$ \pm $0.04 & 4.42$ \pm $0.02 & 4.09$ \pm $0.01 & 4.23$ \pm $0.01 & 3.63$ \pm $0.03 & 3.89$ \pm $0.01 \\
                GPT-4o & 4.29$ \pm $0.04 & 4.42$ \pm $0.03 & 4.19$ \pm $0.03 & 4.21$ \pm $0.00 & 4.28$ \pm $0.02 & 4.47$ \pm $0.02 & 4.20$ \pm $0.01 & 4.23$ \pm $0.02 & 4.10$ \pm $0.02 & 4.24$ \pm $0.01 & \textbf{3.68$ \pm $0.03} & \textbf{4.00$ \pm $0.01} \\
                \hline
                UniChart & 4.56$ \pm $0.00 & 4.65$ \pm $0.01 & 4.47$ \pm $0.01 & 4.57$ \pm $0.01 & 4.39$ \pm $0.01 & 4.50$ \pm $0.01 & 4.24$ \pm $0.00 & 4.39$ \pm $0.01 & N.A. & N.A. & N.A. & N.A. \\
                ChartGemma & 4.60$ \pm $0.00 & 4.68$ \pm $0.00 & 4.53$ \pm $0.01 & 4.62$ \pm $0.01 & \textbf{4.48$ \pm $0.02} & 4.58$ \pm $0.01 & \textbf{4.35$ \pm $0.01} & 4.48$ \pm $0.01 & N.A. & N.A. & N.A. & N.A. \\
                \hline
                MEVER w/o images & 4.61$ \pm $0.01 & 4.68$ \pm $0.02 & 4.54$ \pm $0.02 & 4.62$ \pm $0.01 & 4.41$ \pm $0.02 & 4.52$ \pm $0.01 & 4.25$ \pm $0.02 & 4.40$ \pm $0.03 & 4.12$ \pm $0.01 & 4.19$ \pm $0.01 & 3.47$ \pm $0.02 & 3.78$ \pm $0.02 \\
                \hline
                MEVER (ours) & \textbf{4.64$ \pm $0.03} & \textbf{4.72$ \pm $0.02} & \textbf{4.60$ \pm $0.02} & \textbf{4.67$ \pm $0.03} & \textbf{4.48$ \pm $0.01} & \textbf{4.59$ \pm $0.01} & \textbf{4.35$ \pm $0.01} & \textbf{4.51$ \pm $0.01} & \textbf{4.14$ \pm $0.01} & \textbf{4.25$ \pm $0.01} & 3.66$ \pm $0.01 & 3.97$ \pm $0.02 \\
			\bottomrule
		\end{tabular}
	}
	\label{table:vlm_as_a_judge}
\end{table*}

\subsection{Failure Case and Error Type}

We further examine the generated explanations of both our model and baselines. Here we summarize a main error type. Our model is built upon SciBERT \cite{scibert} and T5 \cite{t5}, which can only capture limited length of evidence texts and explanations. Thus, for extremely long evidence texts and explanations that exceed the maximum length of SciBERT and T5, the quality of explanation generation may be influenced. Similarly, we observe the same performance drop for baseline models, which also have difficulty capturing extremely long sequence. One potential solution is that we can replace our language model with LongFormer \cite{longformer}, which can capture extremely long texts and maintain a good trade-off between model effectiveness and running efficiency. 

\end{document}